\documentclass[12pt]{article}
\usepackage[margin=1in]{geometry}
\usepackage{amsmath,amssymb,amsthm,mathtools}
\usepackage{graphicx}
\usepackage[hidelinks]{hyperref}
\usepackage{xcolor}
\usepackage{enumitem}
\usepackage[capitalise]{cleveref}
\usepackage{booktabs}
\usepackage{makecell}
\usepackage{longtable}
\usepackage{pdfpages}
\usepackage[section]{placeins}

\newtheorem{hypothesis}{Hypothesis}
\newtheorem{interpretation}[hypothesis]{Interpretation}
\newtheorem{remark}[hypothesis]{Remark}
\crefname{hypothesis}{Hypothesis}{Hypotheses}
\Crefname{hypothesis}{Hypothesis}{Hypotheses}
\crefname{interpretation}{Interpretation}{Interpretations}
\Crefname{interpretation}{Interpretation}{Interpretations}
\crefname{remark}{Remark}{Remarks}
\Crefname{remark}{Remark}{Remarks}

\title{Human vs Machine Mathematical Difficulty on Project Euler: An Experimental Analysis}
\author{David Holmes and Johannes Schmitt}
\date{\today}

\begin{document}
\maketitle

\begin{abstract}
We study how the effort and success probability of frontier AI systems scale with human difficulty on problems from Project Euler, an online platform of computational mathematics problems. Our dataset, from the MathArena benchmark \cite{balunovic2025matharena}, consists of $3{,}840$ attempts across 50 problems and 26 model configurations, with problem difficulty measured by the site's public human solve times. Motivated by a proposal of Timothy Gowers, we test a power-law relation $t_{\text{machine}} = a \cdot t_{\text{human}}^b$ between generated-token cost per successful answer and human time, and find $b < 1$ for 20 of the 25 models with usable fits, including the strongest base models; this operationalization therefore does not support an earlier prediction that machines scale worse than humans with difficulty. We also investigate whether success probability on the tested problems can be modeled by a simple exponential decay $p_{\text{success}} = e^{c\,t_{\text{human}}}$, predicting a linear relation between $\log p_{\text{success}}$ and $t_{\text{human}}$. Using a binning approach for data aggregation we find moderate empirical support (median bin-level $R^2 = 0.92$ across the 22 best-covered configurations) for this model. Following METR \cite{kwa2025measuringaiabilitycomplete}, we also fit logistic success curves and extract 50\% task-length horizons $h_{50}$; the strongest configurations in our 20 April 2026 snapshot reach roughly $2.5$--$4.3$~hours on our fastest-five human baseline, with a log-linear fit through the state-of-the-art frontier giving a descriptive doubling time of about $75$~days for the SOTA $h_{50}$.
\end{abstract}

\section{Introduction}

Modern large language models carry an enormous base of mathematical knowledge, and each successive generation can successfully approach harder and harder problems. Yet on sufficiently difficult problems they still fail. A natural question is what mechanism drives these failures as tasks get harder: for instance, does the model's effective search capacity run out, or does it become harder to stay on a correct reasoning path long enough to finish?

At the September 2025 Lorentz Center workshop on Mechanization and Mathematical Research, Timothy Gowers proposed a quantitative picture of the first mechanism, which we take as the starting point for this paper. Consider the task of finding a ``nice'' proof of a mathematical statement of (proof-)length~$n$. In Gowers's framing, a suitably trained research-level mathematician runs a probabilistic proof-search algorithm with expected running time $f(n)$, while a system produced by current AI training methods runs a probabilistic algorithm with expected running time $g(n)$; the hypothesis is that $g$ grows much more rapidly than $f$ as $n$ increases. This is a claim about scaling rather than absolute speed: an AI system may be much faster than any human on small instances and still fall behind on large ones, as long as its expected running time scales badly enough.

Gowers's framing does not commit to specific functional forms for $f$ and $g$. To test it against data one needs a collection of mathematical problems carrying comparable difficulty signals for humans and machines. \href{https://projecteuler.net/}{Project Euler} is a long-running online collection of computational mathematics problems in which participants submit a single numerical answer, obtained through a mix of mathematical insight and programming; for registered users the site records the elapsed time from problem publication to their first correct submission, and publishes the resulting top solver times per problem. This makes it a convenient test bed: public human completion times are available at the per-problem level, and the MathArena project has compiled repeated AI attempts on recent Project Euler problems across many frontier models \cite{matharenaeuler, balunovic2025matharena}. The resulting human and machine measurements are not directly comparable in units, but they give two operational views of difficulty --- how long strong human solvers need, and how much generated computation a model must spend to obtain a correct answer. \Cref{sec:complexitymodel} records the specific power-law instantiation of Gowers's framing we test against this data (Hypothesis~T). \Cref{sec:hypothesis_p_intro} introduces a complementary one-parameter exponential success-probability model (Hypothesis~P), motivated by viewing a long reasoning chain as a process with a roughly constant per-unit-time chance of irrecoverable deviation. \Cref{sec:metr_method} introduces the METR-style logistic task-length model from \cite{kwa2025measuringaiabilitycomplete}, which we also fit to the same data to produce capability horizons.

\subsection{Complexity of Proof Search}
\label{sec:complexitymodel}

Gowers' proposal is an idealized search picture. For a given mathematical problem, humans and machines are both viewed as searching for a proof or solution trajectory, with machines able to do low-level operations much faster, while humans are assumed to prune the search space more effectively.

\begin{hypothesis}[Hypothesis T: Time-Scaling Model]
\label{hyp:time}
The relationship between machine and human solving time is
\begin{equation}
t_{\text{machine}} = a \cdot t_{\text{human}}^b,
\label{eq:time_scaling}
\end{equation}
where $a \ll 1$ reflects faster low-level machine operations and $b > 1$ would indicate that humans scale better with increasing task difficulty.
\end{hypothesis}

If this hypothesis is correct, machines should dominate on easy problems but lose that advantage as human difficulty increases. A methodological caveat that will matter for our empirical test: instead of trying to measure machine wall-clock search time across providers, which might depend on external factors and implementation details, we operationalize $t_{\text{machine}}$ in terms of generated tokens in the API-calls of solution attempts (see \Cref{sec:dataset} for details). The fitted exponent $b$ should be read in that light.

\begin{interpretation}[A Proof-Alphabet Heuristic]
\label[interpretation]{interpretation:alphabet}
Suppose a proof is a string in an alphabet of size $A$, and let $N$ be the effective proof length. A naive search takes on the order of $A^N$ candidate checks. If a search procedure can rule out almost all strings of each fixed length in advance, one may model that by replacing $N$ with an effective length $pN$, where smaller $p$ means stronger pruning. If humans and machines have effective pruning factors $p_{\text{human}}$ and $p_{\text{machine}}$, then one obtains a relation of the form \eqref{eq:time_scaling} with exponent $b = p_{\text{machine}}/p_{\text{human}}$. In that toy model, a value of $b>1$ could e.g.\ be explained if the human search procedure takes larger effective ``intuitive jumps''.
\end{interpretation}

\begin{remark}
This picture is most natural in a setting where candidate proofs can be mechanically verified, for example inside a proof assistant. Project Euler is less formal: each problem asks for a single numerical answer found through a mathematical and computational reduction rather than through a formal proof, and both humans and AI systems may follow incorrect lines of reasoning and never recover. The proof-alphabet framing should therefore be read as an intuitive motivation for Hypothesis~T, not as a literal model of Project Euler solving.
\end{remark}

\subsection{An Exponential Success-Probability Model}
\label{sec:hypothesis_p_intro}

Hypothesis~T is a statement about effort. A complementary question is how the \emph{probability} of producing a correct answer behaves as task difficulty grows. As a null model, suppose a long reasoning trajectory is correct only if it avoids any irrecoverable mistake throughout, and that the chance of such a mistake per unit of work is roughly constant. The probability of success on a task of length $t$ then decays exponentially in $t$:

\begin{hypothesis}[Hypothesis P: Exponential Success Model]
\label{hyp:prob}
For a fixed AI system and task,
\begin{equation}
p_{\text{success}}(\text{model}, \text{task}) = p_{\text{model}}^{\,t_{\text{task}}},
\label{eq:hypothesis_p}
\end{equation}
where $0 < p_{\text{model}} \le 1$ is model-specific and $t_{\text{task}}$ is a task-length proxy.
\end{hypothesis}

We identify $t_{\text{task}}$ with our human-time proxy $t_{\text{human}}$.  As with Hypothesis~T, the constant-deviation-rate picture is meant as a clean motivation rather than a literal description of how AI systems solve Project Euler problems: real reasoning is more structured than a memoryless chain that breaks at random, and Project Euler in particular often rewards a single correct mathematical reduction rather than a long error-free derivation. Even so, the exponential form is the simplest functional shape one could hope to see, and it is empirically interesting how close real models come to it; we examine the empirical fit in \Cref{sec:hypothesis_p_results}.

\subsection{METR-Style Time Horizons}
\label{sec:metr_method}

A third, more functional-form-agnostic way to summarize model capability is to fit success probability directly as a function of human task time. Following METR \cite{kwa2025measuringaiabilitycomplete}, we use a logistic curve:
\begin{equation}
p_{\text{success}}(\text{model}, \text{task})=
\sigma\!\bigl((\log h_{\text{model}}-\log t_{\text{human}}(\text{task}))\beta_{\text{model}}\bigr),
\label{eq:metr}
\end{equation}
where $\sigma$ is the sigmoid function and $\beta_{\text{model}}>0$ controls the steepness of the transition. The parameter $h_{\text{model}}$ is the human task time at which the fitted curve reaches $50\%$ success; we denote it $h_{50}$ below. Because success probability decreases as $t_{\text{human}}$ increases, the $80\%$ horizon $h_{80}$ lies below $h_{50}$. Compared with Hypothesis~P, the METR logistic uses log-time and an additional steepness parameter, so it does not commit to the strict exponential decay shape but pays for that flexibility with an extra free parameter per model.

\section{Related Work}

Our horizon analysis directly follows the general methodology of METR's task-length paper \cite{kwa2025measuringaiabilitycomplete}. We do not import their task-family weighting, because here the units of analysis are individual Project Euler problems rather than heterogeneous benchmark families.

The Hypothesis~P picture --- exponential decay of success probability in task length --- is closely related to multiplicative-difficulty effects in the wider literature on mathematical reasoning. \cite{shah2025aiassistedgenerationdifficultmath} construct MATH$^2$, a benchmark of harder problems built by composing multiple skills drawn from the original MATH dataset \cite{hendrycks2021measuring}, and report that overall solve rates fall approximately as the product of the per-skill solve rates: combining $k$ skills with individual solve rates $p_1, \dots, p_k$ gives a composite solve rate close to $\prod_i p_i$. Hypothesis~P is a coarse time-domain version of the same intuition: extending the reasoning chain by another comparable block multiplies the success probability by a further factor of roughly $p_{\text{model}}^{\Delta t}$, just as adding another skill multiplies it by another per-skill solve rate in the MATH$^2$ setting. \cite{shojaee2025illusionthinkingunderstandingstrengths} study controllable puzzle environments and find sharp performance collapses as complexity increases. Many other papers investigate repeated sampling, pass@k, and related probabilistic views of model reasoning \cite{wang2023selfconsistencyimproveschainthought, lewkowycz2022solving, dekoninck2025openproofcorpuslargescale}.

\section{Experimental Setup}

\subsection{Project Euler Timing Data}

Project Euler publishes new problems according to a fixed schedule. For registered users, the site records the elapsed time between publication and the user's first correct submission. This is not a controlled stopwatch from the moment a user first opens the problem: a solver who starts later may still appear in the top-100 list if the problem is difficult enough.

That detail matters for our human baseline. Later entries among the fastest published solves may partly reflect delayed starts rather than continuous work on the task. We therefore use only the five fastest recorded solves for each problem, whose timestamps typically lie much closer to publication.

\subsection{Dataset Construction}
\label{sec:dataset}

The dataset covers Project Euler Problems 943--992, i.e.\ 50 problems in total. It contains 3{,}840 AI attempts across 26 model configurations. By a \emph{model configuration} we mean a specific base model at a specific reasoning-effort setting, optionally wrapped in a solving loop. The evaluated configurations include base models from OpenAI, Google, Anthropic, xAI, Moonshot, DeepSeek, and Z.ai/GLM, together with several agentic variants for GPT-5 and Grok 4 Fast as described on the MathArena Euler site \cite{matharenaeuler}; MathArena's general methodology is described in \cite{balunovic2025matharena}. Here ``agentic'' means that the base model is embedded in a more elaborate solving loop, for example with iterative decomposition, retries, or tool-using workflow structure, rather than being queried once with a single direct prompt. The Project Euler platform itself is described at \cite{projecteuler}. Our snapshot is the full, non-curated MathArena Euler export available on 2026-04-20. One configuration, GPT-5-Pro (Augmented Solver), is excluded from all fits,\footnote{As documented on the MathArena Euler site \cite{matharenaeuler}, that configuration was evaluated only on the seven problems that had previously gone unsolved by the base models, rather than on a representative cross-section of the 50 problems covered for other configurations. Both its human-time distribution and its overall success rate are therefore conditioned on ``hardest for base models'', so the resulting Hypothesis~T and METR summaries are not directly comparable with the others; we exclude the configuration rather than mix it into the cross-model tables.} so the analyses in this paper report 25 configurations rather than 26.

Coverage is irregular along two separate axes. The first is the set of problems each configuration was evaluated on. Most configurations cover the lower end of the problem range starting at problem 943 (published 2025-05-04), but they end at different problems\footnote{To reduce cost, for any given model the evaluation on newly released Project Euler problems ceases once a new model by the same developer and in the same class is released.} and a few configurations are sparse over their nominal range (e.g.\ GLM 5.1 misses problem 990, and one of the older Grok 4 Fast Euler-agent variants starts at problem 944).
The widest windows go all the way up to problem 992 (published 2026-04-11), held by GPT-5.4 (xhigh), Gemini 3.1 Pro Preview and GLM 5.1. Some older base models were evaluated only up to earlier releases -- for instance Gemini 2.5 Pro and GPT-5 (high) end at problem 970 (2025-11-16) and Grok 4 at problem 973 (2025-12-06). All agentic Euler scaffolds, including the Grok 4 Fast variants and GPT-5 (Euler Agent), correspond to a single one-off MathArena experiment \cite{matharenaeuler} whose batch ends at problem 966 (2025-10-25) and so covers at most 24 problems. The second axis is the number of attempts per covered problem: several base models are evaluated with four attempts per problem, while some agentic variants have one attempt per problem and others have substantially larger attempt counts. Rather than force the dataset into a balanced panel, we keep the empirical number of attempts actually present for each model-problem pair and aggregate success counts accordingly. For each configuration the \textbf{Included} denominator in \Cref{tab:hypothesis_t_summary} is the number of covered problems under this convention, and the \textbf{Problems} column in \Cref{tab:metr_summary} is the same count.

For machine effort we use the dataset's recorded generated-token counts, which include both the visible output tokens and any internal reasoning tokens reported by the provider. For Hypothesis~T we define, at the problem-by-model level,
\begin{equation}
t_{\text{machine}}(\text{task}) = \frac{\text{total generated tokens across all attempts on this task}}{\text{number of successful attempts on this task}},
\label{eq:t_machine}
\end{equation}
restricted to tasks with at least one successful attempt. This is the expected number of generated tokens paid per successful answer if one restarts the model from scratch until a success occurs. We deliberately do \emph{not} use the mean token count over successful attempts alone. That alternative would e.g. reward a guessing strategy that commits to a single high-level plan near the start of each run: if the plan is right the run terminates quickly and is counted; if the plan is wrong the run terminates in failure and is discarded. Such a strategy could in principle look arbitrarily efficient by this alternative metric, while paying no penalty at all for its failed runs. The definition in~\eqref{eq:t_machine} charges every generated token, failed or successful, and divides only by what the user actually receives. Empirically this choice matters: across the full dataset, the median failed attempt uses roughly $2.5\times$ as many generated tokens as the median successful attempt, so crediting only the successful runs would materially understate the token cost of getting an answer.

\subsection{Human-Difficulty Baseline}

For each problem we define $t_{\text{human}}$ as the geometric mean of the five fastest recorded human solve times. The motivation is heuristic but important: because Project Euler times are measured from publication rather than from first exposure, very late entries in the public leaderboard are less reliable as proxies for actual continuous work time. Restricting to the fastest few solvers preferentially samples users who likely began soon after publication.

This convention has an obvious caveat. We are not timing a pre-registered fixed panel of five humans to completion. Instead we observe the left tail of the solve-time distribution and summarize that tail. The resulting $t_{\text{human}}$ should therefore be understood as a transparent proxy for strong-human task time, and it is likely biased downward relative to the mean time that a broader pool of strong solvers would need.

\subsection{Data Processing and Fitting}
\label{sec:data_processing}

For Hypothesis~T we fit \eqref{eq:time_scaling} model-by-model using all covered problems for that model, but only after excluding model-problem pairs whose empirical success rate is below $50\%$. This keeps the token-per-success statistic in~\eqref{eq:t_machine} from being dominated by tasks where almost every attempt is a failure and a single lucky success sets the denominator. For repeated-attempt configurations the filter keeps cells on which the model is at least moderately reliable; for one-attempt-per-problem configurations the filter degenerates to keeping problems where the single attempt succeeded. In both cases, the fitted exponent $b$ should therefore be read as a scaling law for the token cost of obtaining correct answers on solvable cells, not as a measurement of the total search effort the model would spend over the full task distribution, and in particular not over the harder problems where it almost never succeeds. The sensitivity of $b$ to this threshold choice is examined in \Cref{rem:threshold_sensitivity}, once the headline numbers are in hand.

\paragraph{Fitting procedures.} The power law \eqref{eq:time_scaling} is fit model-by-model by ordinary least squares in log-log space with unit weights on the included cells, so the reported $R^2$ is the log-space coefficient of determination from that regression. The METR logistic \eqref{eq:metr} is fit by maximum likelihood on the per-problem binomial counts, so problems with more attempts contribute more to the likelihood automatically; the optimization uses L-BFGS-B with $\beta_{\text{model}}$ constrained to be strictly positive.

To quantify how sensitive the fitted summaries are to the specific problems in the dataset, we attach a nonparametric bootstrap uncertainty interval to each of $b$ in \Cref{tab:hypothesis_t_summary} and $h_{50}$, $h_{80}$ in \Cref{tab:metr_summary}, following the standard recipe in \cite[Ch.~13]{efron1993introduction}. For a given model, let $\mathcal{P}$ denote the set of problems entering its fit (for Hypothesis~T those are the problem cells passing the $50\%$ success filter; for METR those are all covered problems). We draw $B = 10^3$ bootstrap samples of size $|\mathcal{P}|$ from $\mathcal{P}$ with replacement, refit the corresponding estimator on each sample, discard samples on which the refit does not converge to a finite estimate (which can happen for the METR logistic on small problem counts), and report the $5$th and $95$th percentiles of the surviving estimates as a $90\%$ percentile interval. Intuitively, these intervals describe how much the summary would shift if the same model were evaluated on a slightly different set of problems of the same size and the same difficulty distribution; they do not capture other uncertainties such as noise in provider-reported token counts or in the human baseline. All analysis scripts are included in the repository accompanying this paper.

\section{Results}

\begin{figure}[!htbp]
\centering
\includegraphics[width=\textwidth]{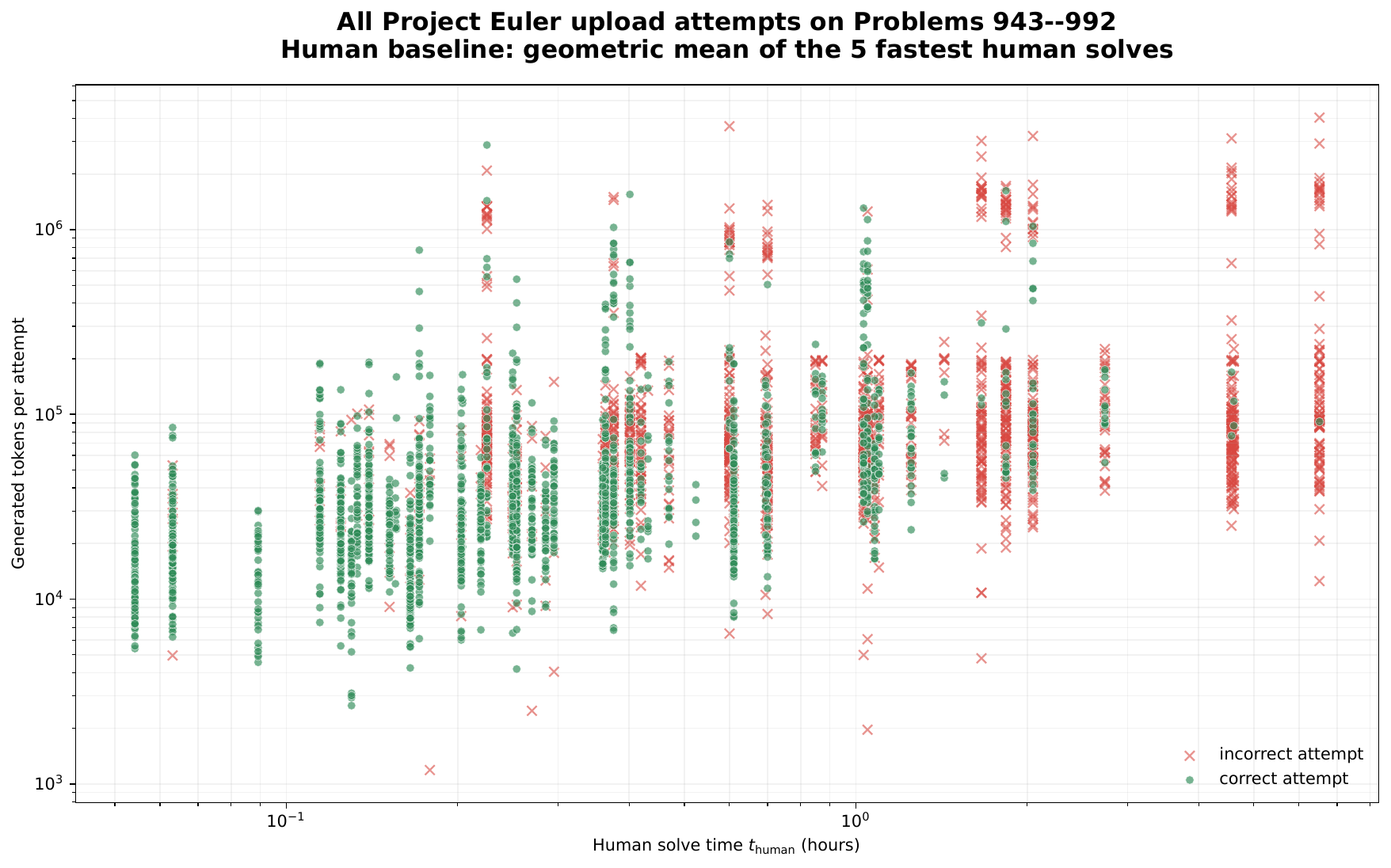}
\caption[All attempts: generated tokens vs.\ human solve time.]{The $3{,}839$ of the $3{,}840$ attempts with a nonzero recorded output-token count, plotted as generated tokens against the fastest-five human solve-time baseline. Green circles are correct attempts; red crosses are incorrect. On easy problems most attempts succeed with a small token count; as $t_{\text{human}}$ increases, the density of red crosses rises and the successful-attempt envelope drifts upward. Visually, no single power law captures the whole cloud, motivating the separate Hypothesis~T and METR summaries that follow.}
\label{fig:all_attempts_scatter}
\end{figure}

Before turning to the fitted summaries, \cref{fig:all_attempts_scatter} shows every individual attempt in the dataset, with colour and marker distinguishing successes from failures. The two qualitative patterns that motivate the rest of the paper are already visible: as human solve time increases (measure of difficulty), failures become more common and generated-token counts trend upward.

\subsection{Hypothesis T: Token Consumption}
\label{sec:hypothesis_t_results}

\begin{figure}[!htbp]
\centering
\includegraphics[width=0.85\textwidth]{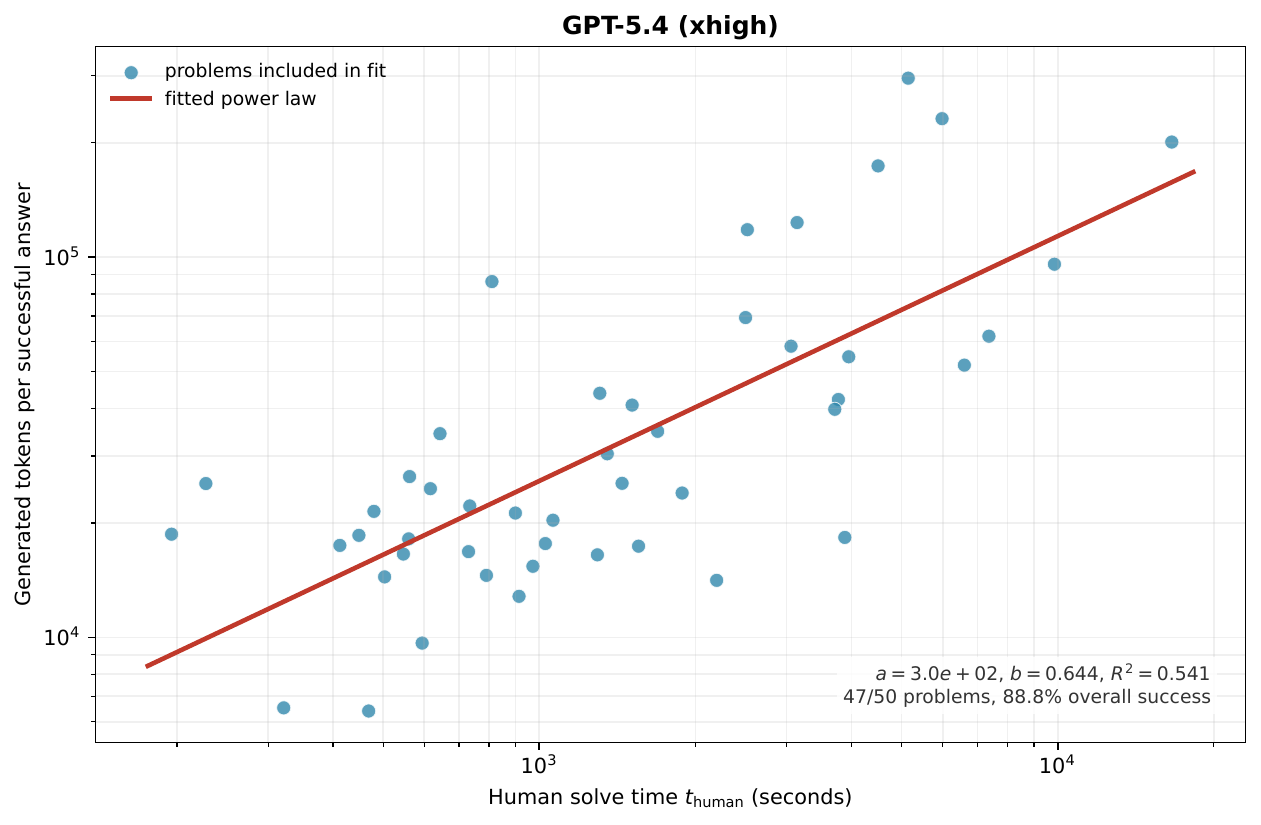}
\caption{Fitted Hypothesis~T power law for GPT-5.4 (xhigh): generated tokens per successful answer plotted against the fastest-five human solve-time baseline (seconds, log-log axes), restricted to the model-problem pairs with empirical success rate at least $50\%$. The fitted exponent is $b = 0.644$ with $R^2 = 0.541$.}
\label{fig:hypothesis_t_gpt54}
\end{figure}

Hypothesis~T predicts that generated-token cost per successful answer should scale superlinearly with human difficulty. \Cref{fig:hypothesis_t_gpt54} shows the fitted log-log plot for GPT-5.4 (xhigh), the best-covered configuration in our dataset. \Cref{app:hypothesis_t_plots} contains the analogous per-model plots for the remaining configurations and the full fitted-parameter table (\Cref{tab:hypothesis_t_summary}).

For many models, especially the stronger and better-covered ones, the fitted relation is at least moderately coherent: $R^2$ values are often in the $0.4$--$0.6$ range, and the log-log plots in \Cref{app:hypothesis_t_plots} do show an interpretable trend. What is unexpected is the direction of that trend: of the $25$ models with usable fits, $20$ have $b < 1$, including the strongest high-coverage base models: Claude Opus 4.6 ($b=0.722$), GPT-5.4 (xhigh) ($b=0.644$), Gemini 3.1 Pro Preview ($b=0.467$), and GPT-5.2 (high) ($b=0.813$). Only five fits give $b>1$, and all five are agentic variants, with one of them---GPT-5 (Euler Agent 2)---based on only four included problems.

Bootstrap uncertainty, reported in the $b$~$90\%$~CI column of \Cref{tab:hypothesis_t_summary}, follows the same pattern. The $90\%$ CIs for three of the four strongest high-coverage base models --- Claude Opus 4.6 ($[0.59, 0.88]$), GPT-5.4 (xhigh) ($[0.49, 0.80]$), and Gemini 3.1 Pro Preview ($[0.36, 0.60]$) --- lie entirely below $1$; GPT-5.2 (high) at $[0.61, 1.03]$ just barely crosses $1$ at the upper end. Conversely, four of the five $b>1$ fits have CIs that either cross $1$ or are extremely wide, reflecting the small number of included problems in those scaffolds: GPT-5 (Euler Agent 2) has $b=1.60$ with CI $[0.36, 7.68]$ on only $4$ cells.

\begin{remark}[Sensitivity to the $50\%$ filter]
\label[remark]{rem:threshold_sensitivity}
The choice of the $50\%$ threshold for including data points (as explained in \Cref{sec:data_processing}) does not drive the main $b<1$ finding. For the three strongest high-coverage base models, rerunning the fit across thresholds $\{0.25, 0.5, 0.75\}$ keeps $b$ below $1$ throughout: GPT-5.4 (xhigh) moves from $b=0.752$ at threshold $0.25$ to $b=0.510$ at threshold $0.75$; Claude Opus 4.6 from $0.773$ to $0.674$; and Gemini 3.1 Pro Preview from $0.547$ to $0.425$. The lower end of this range admits cells with noisier single-success denominators and tends to pull $b$ upward; the higher end restricts to the easiest cells and pulls $b$ downward. Neither end produces $b\ge 1$ on these models. 
We cannot set the threshold to 0 as this would yield infinite values in \eqref{eq:t_machine} for problems where none of the model attempts was successful. This means that we are effectively ignoring the fact that some (human-solvable) problems are simply too hard for these models, despite their superior scaling indicated by $b < 1$. We take up this theme in Section \ref{sec:speculation}. 
\end{remark}

So in this operationalization, $b < 1$ says that the effort ratio between machines and humans actually \emph{improves} in favor of the machine as tasks become harder. Within the proof-search framing of \Cref{sec:complexitymodel}, this is the opposite of the prediction Hypothesis~T was designed to test: the AI systems we evaluated appear to scale \emph{better} than the strong human Project Euler solvers as problem difficulty grows. We discuss the structural caveats in \Cref{sec:interpreting_hyp_t}, but on this benchmark and with this operationalization of $t_{\text{machine}}$ the directional finding is robust to our sensitivity checks (\Cref{rem:threshold_sensitivity}), so we report it as such rather than treating $b < 1$ as an artifact to be explained away.

\subsection{Hypothesis P: Success Probability Decay}
\label{sec:hypothesis_p_results}

Hypothesis~P is a one-parameter model once the time unit is fixed (the only free parameter is $p_{\text{model}}$), and predicts a linear relation between $\log p_{\text{success}}$ and $t_{\text{human}}$ passing through $(0, 0)$ in those coordinates. To fit it model-by-model, we sort the covered problems by $t_{\text{human}}$, partition them into six equal-size quantile bins, compute the mean success rate per bin, and perform a least-squares fit of the bin averages to
\begin{equation}
\log p_{\text{success}} = c \cdot t_{\text{human}}, \qquad c = \log p_{\text{model}} < 0,
\label{eq:hyp_p_linear}
\end{equation}
with the line constrained to pass through $(t_{\text{human}}, p_{\text{success}}) = (0, 1)$. Bins with $0\%$ empirical success are dropped from the regression rather than replaced by an arbitrary floor. We summarize each fit by its bin-level $R^2$ and by the corresponding half-life $-\log 2 / c$, which is the human work-time at which the fitted curve falls to $50\%$ success.

\begin{figure}[!htbp]
\centering
\includegraphics[width=0.85\textwidth]{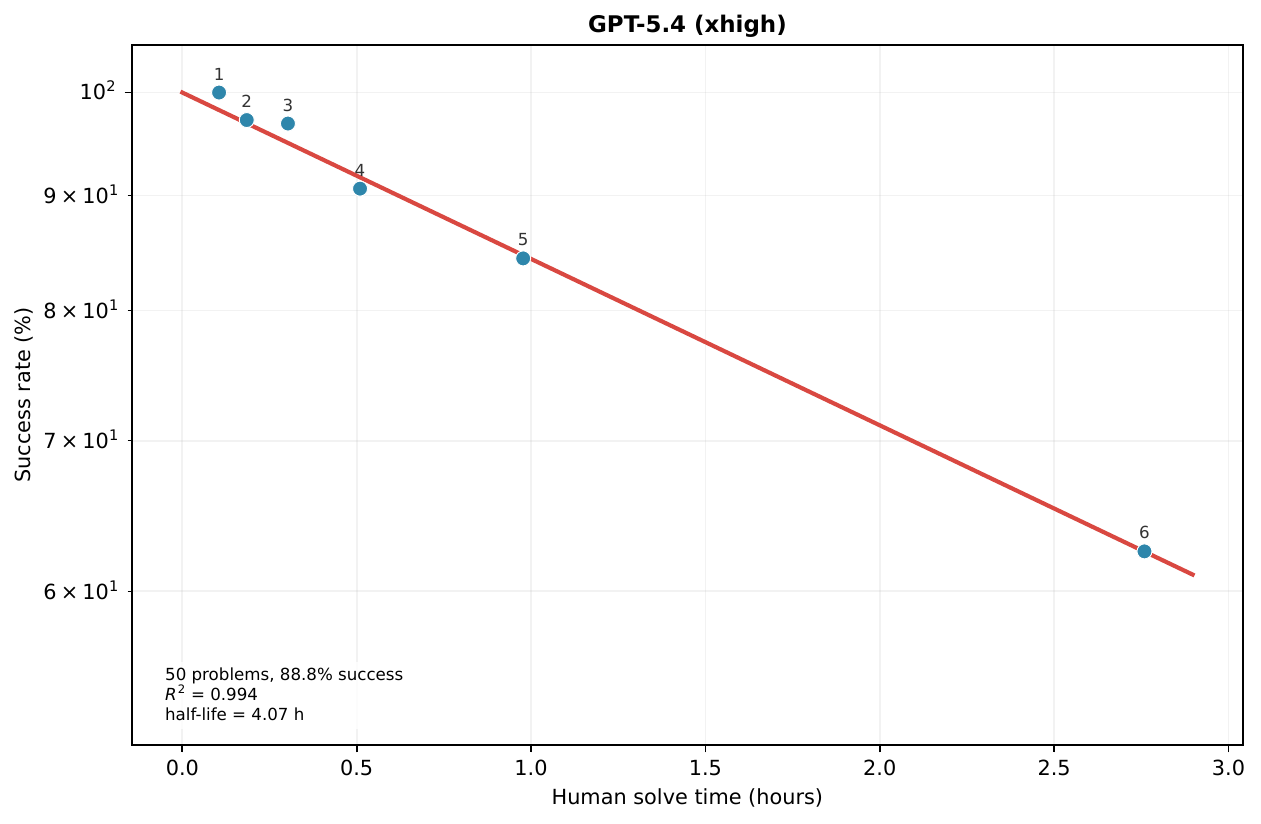}
\caption{Hypothesis~P semi-log exponential-fit panel for GPT-5.4 (xhigh). Bins are equal-size quantile bins of the covered problems sorted by $t_{\text{human}}$, and the line is constrained to pass through $(t_{\text{human}}, p_{\text{success}}) = (0, 1)$. Numbers next to the points are bin indices.}
\label{fig:hypothesis_p_gpt54}
\end{figure}

\Cref{fig:hypothesis_p_gpt54} shows the fit for GPT-5.4 (xhigh), the best-covered configuration in our dataset. \Cref{app:hypothesis_p} contains the analogous panels for the other five most-covered models (\Cref{fig:hypothesis_p_top6}) and the full 22-model fitted-parameter table (\Cref{tab:hypothesis_p_summary}).

The bin-level fit quality is moderately good for many configurations. The median bin-level $R^2$ across the 22 models is $0.92$, with 13 of 22 models exceeding $R^2 = 0.90$ and 9 of 22 exceeding $R^2 = 0.95$. Several of the strongest base models sit very close to the fitted line: GPT-5.4 (xhigh) at $R^2 = 0.994$, GLM 5.1 at $0.971$, Gemini 3.1 Pro Preview at $0.954$, Kimi K2.5 (Think) at $0.949$, and Gemini 3 Flash at $0.930$. Several Grok 4 Fast Euler-Agent variants reach $R^2$ between $0.97$ and $0.99$, and o4-mini (high) reaches $0.967$. The clearest underperformers are GPT-5.1 (high) at $R^2 = 0.472$, GPT-5 (high) at $0.650$, and Grok 4 at $0.703$. The fitted half-lives span a roughly $34\times$ range across the 22 models, from $0.12$~hours (Gemini 2.5 Pro) to $4.07$~hours (GPT-5.4 (xhigh)), and they track the METR-style $h_{50}$ horizons closely, as one would expect from two related parametric fits to the same data; we therefore read Hypothesis~P as both an empirical adequacy check on the exponential shape and a complementary one-parameter summary of the success-vs-difficulty trend.

\begin{remark}[Two-parameter sanity check]
\label[remark]{rem:hypp_two_param}
If we relax \eqref{eq:hyp_p_linear} to a free-intercept fit $\log p_{\text{success}} = \alpha + c \cdot t_{\text{human}}$ on the same bins, the fitted intercepts $y(0\,\text{h}) = e^{\alpha}$ cluster around the $100\%$ value that the strict one-parameter form would have predicted. Across the 22 models the median fitted intercept is $106\%$ and 18 of 22 lie in $[85\%, 115\%]$, with the four outliers being Kimi K2 Thinking ($159\%$), o4-mini (high) ($125\%$), Gemini 2.5 Pro ($80\%$), and the Grok 4 Fast (August solver scaffold) ($77\%$). The intercepts being close to $100\%$ is partly --- but not wholly --- mechanical: for the strongest base models the smallest bin already has a per-bin success rate close to $100\%$ at small $t_{\text{human}}$, so the regression's extrapolation toward $t = 0$ does not have far to travel. With that caveat, the cluster is mild evidence that the constrained one-parameter form is not too far off the unconstrained fit.
\end{remark}

The quantitative summaries are visibly sensitive to the bin count, which we examine in \Cref{app:hypothesis_p}: for GPT-5.4 (xhigh) under the one-parameter fit, the fitted half-life moves from $3.76$~hours at 5 bins to $5.01$~hours at 8 bins, and bin-level $R^2$ stays above $0.98$ for $\{4, 5, 6\}$ bins but drops to $0.85$--$0.89$ for $\{7, 8\}$ bins as per-bin sample size shrinks. We use 6 bins in the main text as a compromise between resolution along $t_{\text{human}}$ and per-bin stability.

\subsection{METR-Style Project Euler Horizons}
\label{sec:metr_results}

\begin{figure}[!htbp]
\centering
\includegraphics[width=\textwidth]{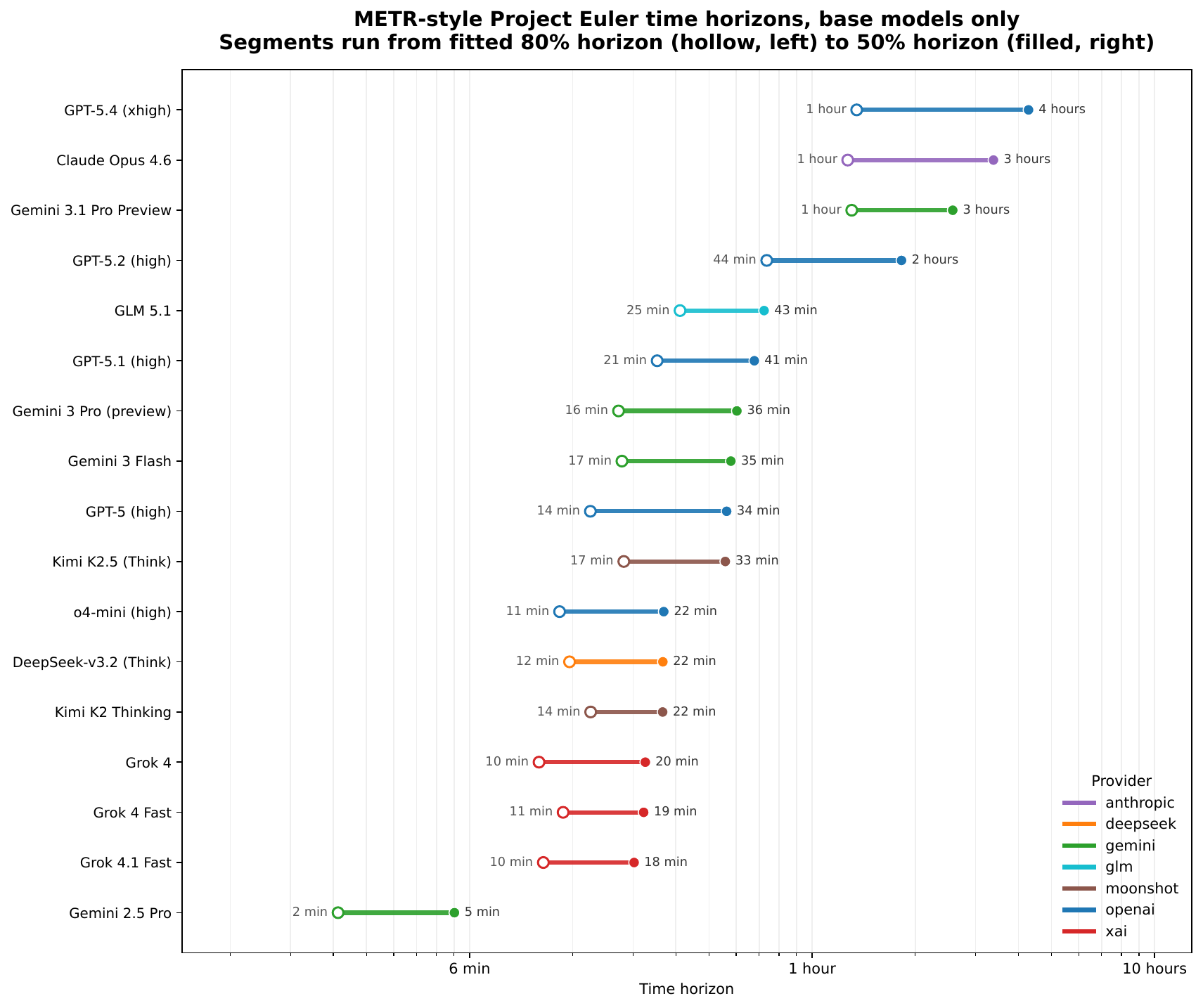}
\caption{METR-style Project Euler horizons for base-model configurations (agentic scaffolds excluded). Each segment runs from the fitted $80\%$ horizon on the left to the fitted $50\%$ horizon on the right; the numerical value of each endpoint is printed next to it in human-readable units. Colors identify the model provider.}
\label{fig:metr_horizons}
\end{figure}

\Cref{fig:metr_horizons} summarizes the METR-style horizons for the base-model configurations, ranked by fitted $h_{50}$. Each horizontal segment runs from the model's $80\%$ horizon (hollow endpoint) to its $50\%$ horizon (filled endpoint), both on a log-hours axis.

The strongest $50\%$ horizons in this dataset are achieved by GPT-5.4 (xhigh) at $4.28$ hours, Claude Opus 4.6 at $3.38$ hours, and Gemini 3.1 Pro Preview at $2.57$ hours. The $h_{50}$ values for these models lie in a sparse part of the empirical $t_{\text{human}}$ distribution --- only $3$--$4$ of the $50$ covered problems have $t_{\text{human}}$ above $h_{50}$ for any of these models --- so they are anchored by relatively few problems and should be read as model-based summaries rather than directly observed crossing points. The fitted $80\%$ horizons sit at $1$--$1.35$~hours and are anchored by many more problems. Bootstrap $90\%$ CIs for both horizons are listed in \Cref{tab:metr_summary}; for the strongest models the $h_{50}$ CIs are still fairly wide (for instance Claude Opus 4.6's $h_{50}$ CI is $[1.77, 9.25]$~hours), so the precise rank-ordering among the top models should not be read too sharply.

Agentic wrappers can substantially extend these horizons, even when the underlying base model and its release date are held fixed. \Cref{tab:metr_summary} includes two such comparison pairs: GPT-5 (high) at $h_{50}=0.56$~hours versus GPT-5 (Euler Agent) at $1.44$~hours, and Grok 4 Fast at $0.32$~hours versus Grok 4 Fast (Euler Agent 1) at $1.73$~hours. In the available configurations, these wrappers are associated with substantial $h_{50}$ gains relative to the base models they are built on. This is not a clean controlled comparison (attempt counts, scaffolding details, and retry/tool-use structure vary between the base and agentic entries), so we read it as an observational fact about the benchmark rather than as evidence that any particular workflow change causes the horizon increase.

\begin{figure}[!htbp]
\centering
\includegraphics[width=\textwidth]{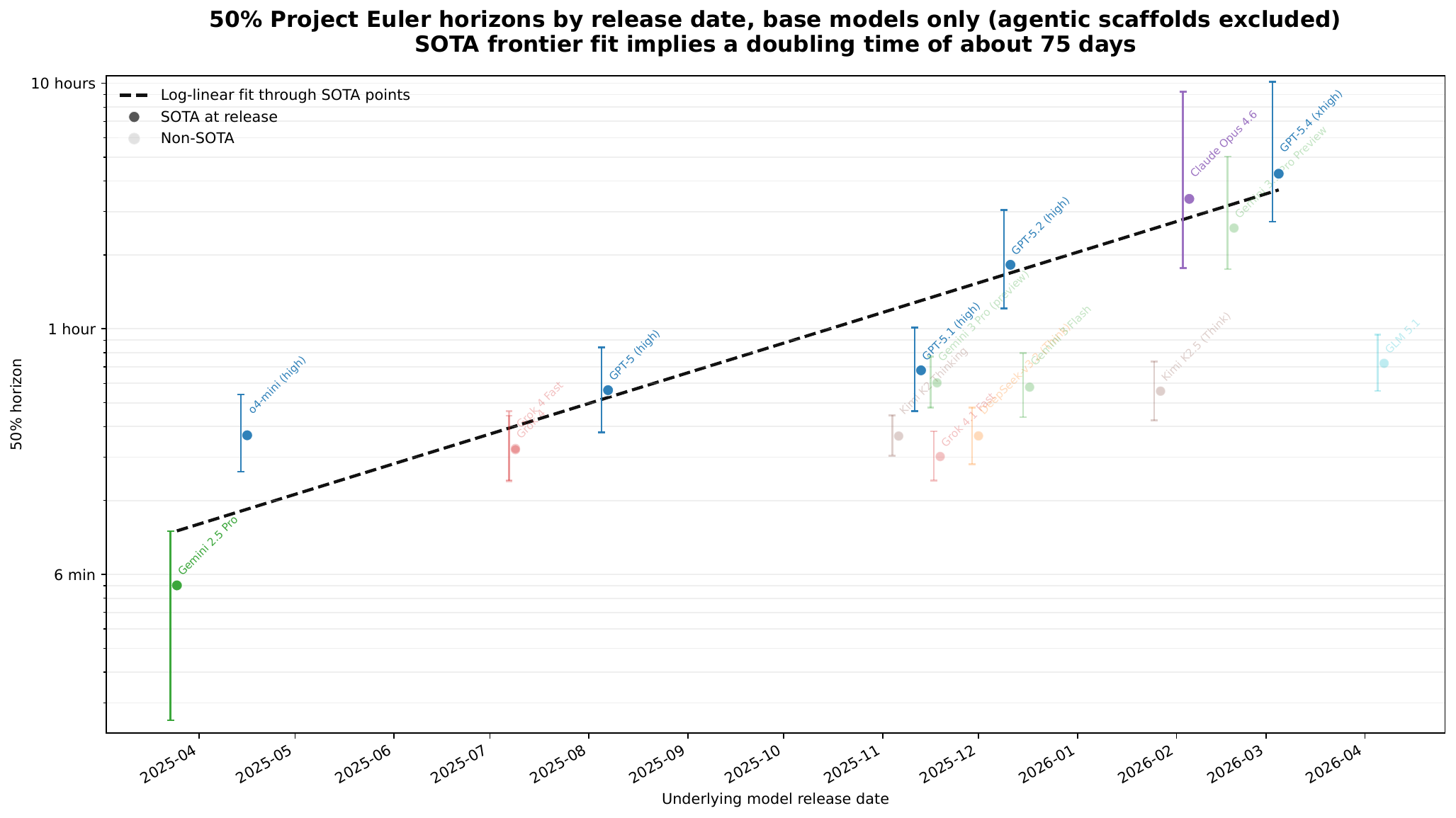}
\caption{$50\%$ Project Euler horizons by release date for base-model configurations (agentic scaffolds excluded). Solid points are SOTA at release; faded points are non-SOTA at their release date. The dashed line is a log-linear regression through the SOTA points and corresponds to a doubling time of approximately $75$~days. Whiskers show bootstrap $5$--$95\%$ intervals for $h_{50}$.}
\label{fig:metr_sota}
\end{figure}

To isolate the underlying base-model trend over time, \Cref{fig:metr_sota} plots $h_{50}$ against release date and fits a log-linear trajectory through the state-of-the-art frontier --- the subset of releases whose $50\%$ horizon exceeded that of every earlier base model. Over the $\approx 345$-day span covered by this dataset, the $h_{50}$ frontier grows from $0.09$~hours (Gemini 2.5 Pro, 2025-03-25) to $4.28$~hours (GPT-5.4 (xhigh), 2026-03-05), a $47\times$ increase, and the log-linear fit through the seven SOTA points implies a doubling time of roughly $75$~days. We report this as a compact descriptive statistic rather than a stable law of progress: the regression rests on only seven points on a log scale, and each individual $h_{50}$ is sensitive to the human-time baseline (see \Cref{tab:top_k_sensitivity}). It is directly analogous to the doubling-time summary reported by METR on their general task-length benchmark \cite{kwa2025measuringaiabilitycomplete}, restricted here to Project Euler under the fastest-five baseline.

\Cref{app:metr_plots} contains the full METR fitted-parameter table (\Cref{tab:metr_summary}, with bootstrap CIs on $h_{50}$ and $h_{80}$) and the per-model logistic panels.

\section{Discussion and Future Directions}

\subsection{Interpreting Hypothesis T}
\label{sec:interpreting_hyp_t}

Taken at face value, the fitted exponents $b < 1$ say that the AI systems we evaluated scale \emph{better} than the strong human solvers in this Project Euler dataset as problems get harder: the cost ratio $t_{\text{machine}} / t_{\text{human}}$ shrinks rather than grows with $t_{\text{human}}$. The most immediate structural worry, that the $50\%$ success-rate filter used to define $t_{\text{machine}}$ on solvable cells is itself responsible for the pattern, does not survive \Cref{rem:threshold_sensitivity}: $b$ stays below~$1$ across thresholds $\{0.25, 0.5, 0.75\}$ for the three strongest high-coverage base models. A separate question is why Project Euler in particular is a benchmark on which this advantage shows up cleanly; we take that up below in \Cref{sec:speculation}.

The agentic variants offer a soft consistency check: several Grok 4 Fast wrappers are among the few configurations with $b > 1$, which is at least compatible with the picture that explicit multi-step workflows expose some of the search cost in the token count (via iterative decomposition or tool use) and shift scaling toward the human-like regime. We do not push this further, since the exponents vary across variants and the included-problem counts for these scaffolds are small. Still, directionally this points towards elaborate agent harnesses leading to extended search behavior by the models, which (within our limited data) did scale worse than the human response times with question difficulty.

\subsection{Interpreting Hypothesis P}
\label{sec:interpreting_hyp_p}

The Hypothesis~P fit is a more positive empirical check than Hypothesis~T: for most well-covered models the binned success rates sit close to the constrained line $\log p_{\text{success}} = c \cdot t_{\text{human}}$ in semi-log coordinates, and the fitted decay rates produce sensible orderings of the strongest base configurations. The bin count is small (six points per fit), so a single high-$R^2$ panel is suggestive rather than definitive evidence for the functional form. The breadth of the pattern --- $R^2 > 0.95$ across nearly half of the 22 configurations and $R^2 > 0.90$ across the majority --- is harder to dismiss as coincidence than any single panel would be, and we read the exponential shape as a reasonable first-order description for the bulk of the dataset. \Cref{rem:hypp_two_param} adds an additional check: when we relax the constraint and fit a free intercept, the fitted intercept lands close to the model-implied $100\%$ for most configurations, even though the data does not include a bin at $t = 0$ to anchor it.

The fits are heterogeneous. GPT-5.1 (high), GPT-5 (high), and Grok 4 each have weaker bin-level $R^2$, and we do not have a clean account of why these specific configurations resist the exponential fit while their siblings (GPT-5.2 (high), GPT-5.4 (xhigh), Grok 4.1 Fast) match it well. \Cref{app:hypothesis_p} additionally shows that for GPT-5.4 (xhigh) the fitted half-life moves over a range of about $1.3$~hours with the choice of bin count even when the fit quality stays high, so $p_{\text{model}}$ is best read as a descriptive summary rather than as a calibrated parameter.

A more substantive limitation of Hypothesis~P at this benchmark is that Project Euler is not the cleanest test bed for its underlying picture. A Project Euler problem typically rewards finding a single mathematical reduction; once that reduction is found, there is usually not much remaining length over which a deviation could occur. The constant-deviation framing is more natural for problems whose solution is a long chain of comparable inferential steps, such as multi-step proofs. We therefore read the empirical good fit on Project Euler as encouraging but not decisive evidence for that picture.

\subsection{Interpreting METR-Style Horizons}

The METR methodology is a useful and robust descriptive tool for this kind of data. It uses all covered problems directly and produces a natural pair of horizons ($h_{50}$ and $h_{80}$) that are easy to read as ``the human work-time at which the model reaches $50\%$ or $80\%$ success''. On the current Project Euler dataset these horizons make base-model rankings and base/agent comparisons directly readable from \Cref{fig:metr_horizons} and \Cref{tab:metr_summary}, and the release-date trend of the base-model state-of-the-art from \Cref{fig:metr_sota}.

What METR does not provide is a causal interpretation or an underlying model of solving. The logistic curve in \eqref{eq:metr} is a convenient functional form rather than a statement about how a model's reasoning actually degrades with task difficulty. By comparison, Hypothesis~P (\Cref{sec:hypothesis_p_results}) takes a more committed stance on the functional form --- exponential decay in $t_{\text{human}}$ rather than logistic decay in $\log t_{\text{human}}$ --- but on this dataset the two analyses agree closely at the level of model rankings and produce comparable order-of-magnitude horizons. The last column of \Cref{tab:hypothesis_p_summary} reports the METR $h_{50}$ alongside Hypothesis~P's $t_{50}$ for direct comparison.

The fitted $h_{50}$ and $h_{80}$ should be read as compact, benchmark-specific success-vs-difficulty summaries rather than measurements of a mechanism or universal model constants: they depend on our choice of human baseline, on the task family, and on the empirical mix of attempts available for each model.

\subsection{Why is $b < 1$? Some speculation}
\label{sec:speculation}

Within the proof-search framing of \Cref{sec:complexitymodel}, the Hypothesis~T fits with $b < 1$ suggest that current frontier systems already scale better than strong human solvers on Project Euler problems. This sits in tension with their failure to solve the very hardest Project Euler problems, and more generally with the absence of comparable AI contributions to research mathematics: why do systems that beat strong humans on most problems of a public computational-mathematics benchmark fail on the very hardest problems, and more importantly have so far failed to visibly accelerate research-level discovery? We close the discussion section with five non-mutually-exclusive speculative explanations; these are conjectures rather than claims supported by our data.

\paragraph{(i) Project Euler problems mix thinking and implementation, and AI is faster at the implementation half.} A Project Euler problem typically combines a mathematical insight (find the right reduction or technique) with an implementation step (write code that actually computes the numerical answer). That implementation step --- nontrivial code, debugging, computational tricks --- can take a strong human a substantial fraction of their total solve time. Current LLMs handle implementation much faster, so even if the mathematical-thinking part scales as Hypothesis~T originally predicted, an AI's total solve time can come out below the human's once the implementation half dominates: a problem on which a human spends $30$~minutes thinking and $2$~hours implementing, while a model spends $45$~minutes thinking and $15$~minutes implementing, looks like $b < 1$ on aggregate time even though the model is genuinely slower on the thinking half. A related but distinct mechanism is that LLMs can in some cases substitute their fast implementation for explicit mathematical reasoning, e.g., by trying candidate strategies through code rather than by deriving the right one. Either way, on a benchmark whose problems are dominated more heavily by mathematical thinking rather than implementation, the picture from Hypothesis~T might look quite different.

\paragraph{(ii) Reasoning degrades over long trajectories.} Hypothesis~P already says that, empirically, success probability decays in $t_{\text{human}}$. A natural mechanistic reading is that current LLMs cannot reliably maintain a coherent reasoning chain over many hours of work: subtle missteps and dead-end excursions accumulate, and the model's self-correction rate is bounded. There may also be a more structural failure mode, sometimes called \emph{context rot}: a long inference creates many model-generated tokens relative to the higher-entropy initial human prompt, so the trajectory drifts further from the well-anchored conversational regime the model was trained on, possibly contributing to increasingly erratic late-trajectory thinking. Research-level mathematics typically demands sustained reasoning over days or weeks, far beyond the single-call horizon of any current frontier system.

\paragraph{(iii) Big toolbox, limited tool-building.} LLMs are pre-trained on a vast library of standard mathematical techniques and PE-style algorithmic patterns; most Project Euler problems can be reduced to one of these standard tools. Strong human PE solvers know fewer tools off-the-shelf but are versatile at improvising and combining them, so their search trees on PE are larger --- consistent with the observed $b < 1$ on this benchmark. Research-level mathematics, by contrast, often requires \emph{inventing} new tools (definitions, frameworks, partial theories) before the problem is even tractable. There is no clear evidence yet that current LLMs are competitive with humans on this kind of creative tool-building.

\paragraph{(iv) Scaffolding and architecture for long-horizon work.} Significant research progress is rarely a single inference: it involves long iterative theory-building, accumulation of problem-specific intuition, and revisiting half-finished arguments over weeks or months. Today's LLMs operate within a single context window, and even sophisticated agentic scaffolds do not yet reproduce a researcher's long-horizon workflow. It is plausible that the right combination of scaffolding, persistent memory, problem-specific finetuning (analogous to a researcher's evolving expertise), or new architectures is simply not yet in place.

\paragraph{(v) ``Overthinking'' on easy problems may flatten the fitted slope.} Another possible explanation for the $b<1$ pattern comes from looking at the easy end of the $t_{\text{human}}$ axis rather than the hard end. There is a growing body of work documenting \emph{overthinking} in reasoning-tuned LLMs: on tasks easy relative to a model's capability, current reasoning models tend to spend disproportionately many tokens on self-checking, alternative-method exploration, and hesitation, with little or no gain in final-answer accuracy and sometimes with measurable accuracy drops \cite{sui2025stopoverthinking, chen2024dontthinkthatmuch, su2025betweenunderoverthinking, srivastava2025llmthinkbench}. A related strand even documents \emph{inverse scaling in test-time compute}, where deliberately lengthening reasoning traces lowers accuracy \cite{gema2025inversescaling}, though that evidence comes from synthetic distractor and puzzle tasks far removed from Project Euler and so bears on our setting only indirectly. Mechanically, an approximately constant token overhead added to every solve produces a much larger relative shift in $\log t_{\text{machine}}$ on easy cells than on hard ones, which would tilt our fitted slope $b$ downward without reflecting better scaling on the underlying mathematical reasoning. The cited overthinking literature is, however, mostly concerned with substantially more trivial tasks (basic arithmetic, GSM8K-style word problems, short text queries) than even the easiest Project Euler problems, so the quantitative size of the effect on our dataset is unclear. The picture also interacts with the asymmetric evaluation affordances and goals discussed in \Cref{sec:limitations}: a model whose explicit objective is producing a correct boxed answer is naturally incentivized to over-verify before committing, in a way that a human optimizing for ``be the first to submit a correct answer'' is not. It would be interesting to rerun the analysis under explicit ``answer as quickly as possible'' prompts, tighter token budgets, or evaluation methodologies that allow the model to commit early, but this lies beyond the scope of the present paper.

\bigskip

These five explanations are not mutually exclusive, and the gap between Project Euler and research mathematics is plausibly some combination of all of them. Pinning down their relative contributions seems like a productive direction for future evaluation work.

\subsection{Limitations}
\label{sec:limitations}

The largest limitation is the human baseline itself. The fastest-five convention is motivated by the publication-time issue, but it samples only the left tail of the human solve-time distribution. To see how strongly this choice matters, \cref{tab:top_k_sensitivity} recomputes the Hypothesis~T exponent and the METR $h_{50}$ horizon for the three strongest high-coverage base models under $t_{\text{human}}$ taken to be the geometric mean of the $k$ fastest solvers for $k \in \{1,3,5,10,20,50\}$. The exponents $b$ are quite stable, moving by around $0.2$ over the full range of $k$; the absolute horizons $h_{50}$ grow monotonically with $k$ because larger $k$ pulls in slower solvers and so stretches the $x$-axis.

\begin{table}[t]
\centering
\small
\caption{Sensitivity of Hypothesis~T's power-law exponent and METR's 50\% horizon to the choice of $k$ in the top-$k$ human-time baseline, for the three strongest high-coverage base models. $b$ is the exponent in \eqref{eq:time_scaling}; $h_{50}$ is the 50\% horizon from \eqref{eq:metr}; $\overline{t_{\text{human}}}$ is the geometric mean of the per-problem $t_{\text{human}}$ values across the 50 problems under that baseline.}
\label{tab:top_k_sensitivity}
\begin{tabular}{lrrrrrr}
\toprule
\textbf{Model} & \textbf{$k$} & \textbf{$b$} & \textbf{$R^2$} & \textbf{$h_{50}$ (h)} & \textbf{$h_{80}$ (h)} & \textbf{$\overline{t_{\text{human}}}$ (h)} \\
\midrule
GPT-5.4 (xhigh)          & 1  & 0.669 & 0.607 & 2.68   & 0.84  & 0.26 \\
GPT-5.4 (xhigh)          & 3  & 0.656 & 0.556 & 3.50   & 1.10  & 0.35 \\
GPT-5.4 (xhigh)          & 5  & 0.644 & 0.541 & 4.28   & 1.35  & 0.43 \\
GPT-5.4 (xhigh)          & 10 & 0.610 & 0.530 & 7.16   & 2.08  & 0.62 \\
GPT-5.4 (xhigh)          & 20 & 0.558 & 0.541 & 17.93  & 4.17  & 1.05 \\
GPT-5.4 (xhigh)          & 50 & 0.472 & 0.548 & 107.12 & 17.35 & 3.25 \\
\midrule
Claude Opus 4.6          & 1  & 0.700 & 0.561 & 2.66   & 0.82  & 0.26 \\
Claude Opus 4.6          & 3  & 0.727 & 0.566 & 3.06   & 1.05  & 0.35 \\
Claude Opus 4.6          & 5  & 0.722 & 0.564 & 3.38   & 1.27  & 0.43 \\
Claude Opus 4.6          & 10 & 0.708 & 0.566 & 4.30   & 1.89  & 0.62 \\
Claude Opus 4.6          & 20 & 0.655 & 0.569 & 8.41   & 3.65  & 1.05 \\
Claude Opus 4.6          & 50 & 0.560 & 0.575 & 37.50  & 14.74 & 3.25 \\
\midrule
Gemini 3.1 Pro Preview   & 1  & 0.450 & 0.475 & 1.73   & 0.82  & 0.26 \\
Gemini 3.1 Pro Preview   & 3  & 0.467 & 0.467 & 2.16   & 1.07  & 0.35 \\
Gemini 3.1 Pro Preview   & 5  & 0.467 & 0.468 & 2.57   & 1.30  & 0.43 \\
Gemini 3.1 Pro Preview   & 10 & 0.457 & 0.475 & 4.03   & 1.97  & 0.62 \\
Gemini 3.1 Pro Preview   & 20 & 0.409 & 0.461 & 9.16   & 3.80  & 1.05 \\
Gemini 3.1 Pro Preview   & 50 & 0.332 & 0.427 & 44.90  & 15.28 & 3.25 \\
\bottomrule
\end{tabular}
\end{table}

The $h_{50}$ dependence has a natural one-sided interpretation. The METR methodology as applied elsewhere \cite{kwa2025measuringaiabilitycomplete} would ideally time a fixed panel of qualified solvers to completion on every task. Our fastest-$k$ proxy samples only the left tail of whatever such a panel would produce, because it keeps the $k$ solvers who finished the fastest. Under the assumption that for each problem at least $k$ qualified solvers began work near publication and worked continuously until submission, this proxy is a lower-tail estimate of strong-human time, and the geometric mean of the top $k$ times would be a lower bound on the geometric-mean time of a broader fixed panel, so that the reported $h_{50}$ and $h_{80}$ would be lower bounds on the METR-ideal horizons. The Project Euler timestamps are not controlled measurements and this assumption is substantial, so we state the implication rather than treat it as a certified bound. Concretely, the GPT-5.4 (xhigh) fastest-$1$ baseline gives $h_{50}=2.68$~hours, which is the strongest statement we can make about that model from this dataset: even on the most generous reading of $t_{\text{human}}$, GPT-5.4 (xhigh) still hits the 50\% success floor only at roughly $2.7$~hours of human work. Our preferred value of $k=5$ is chosen as a middle ground: $k=1$ is very noisy, and $k$ much above ten inevitably includes solvers who began the problem hours or days after publication and therefore have inflated elapsed-time numbers.

A separate concern about the human baseline is that AI tools have become widely available over the publication window of our covered Project Euler problems (943--992, published 2025-05-04 to 2026-04-11), so some of the recorded fastest solves may themselves reflect human--AI collaboration rather than purely human effort, especially toward the more recent problems.

A second limitation is that the human and AI evaluation procedures behind our two operational difficulty proxies are not directly symmetric. Project Euler scores a registered human by the elapsed wall-clock time from publication to their \emph{first correct} submission, with no explicit penalty for incorrect submissions: the implicit incentive is essentially to be the first to submit a correct answer, and the human can iterate freely on intermediate guesses against the site's automatic verifier. The MathArena evaluations \cite{balunovic2025matharena} score AI configurations on whether they reach a correct boxed answer within a bounded budget of code-execution tool calls (200 calls per attempt in the published Euler config); in the non-agentic configurations the model produces a single final boxed answer at the end of one such attempt, with no in-run feedback on whether that answer is correct, so its explicit objective is producing a correct boxed answer within the tool-call budget rather than minimizing wall-clock or token cost. The agentic Euler scaffolds add further structure on top --- explicit decomposition, retries, or verifier interactions --- letting them submit several candidate answers and receive ``correct''/``wrong'' feedback within a single attempt. None of our analyses correct for these structural differences; they should be kept in mind when reading the operational comparison between $t_{\text{human}}$ and $t_{\text{machine}}$.

A third limitation is that provider-reported generated-token counts are not perfectly standardized across model families. We use them because they are the only machine-effort measure available consistently across the dataset, but cross-provider comparisons should still be interpreted cautiously. Generated-token count is moreover only a surface proxy for effective reasoning effort: \cite{chen2026thinkdeep} find that length can correlate weakly or even negatively with correctness and propose a depth-based alternative, but computing it requires access to a model's internal layer-wise states, which the closed cross-provider APIs in our dataset do not expose. A related caveat is that many model configurations operate under hard output-token caps; an attempt that would otherwise continue past the cap is recorded as a failure. Failed-run tokens still enter the numerator of $t_{\text{machine}}$, so the cap does not directly erase their cost on the cells that remain in the fit. The remaining indirect concern --- that the cap pushes the hardest problems below the $50\%$ filter used in \Cref{sec:data_processing} and so removes them from the right of the $t_{\text{human}}$ axis --- also does not appear to drive the $b < 1$ pattern, since GPT-5.4 (xhigh) has $b = 0.752$ at the $25\%$ filter and $b=0.510$ at the $75\%$ filter (\Cref{rem:threshold_sensitivity}).%

Finally, Project Euler measures a specific kind of mathematical problem solving: numerical answers to self-contained computational math problems. It is informative precisely because it is concrete and auditable, but it is not identical to formal theorem proving or to broader research mathematics.

\section{Conclusion}
\label{sec:conclusion}

On this Project Euler dataset, the time-scaling picture behind Hypothesis~T is reversed from the original prediction. Restricted to the model-problem pairs where the model is already moderately reliable, a power-law relation between generated-token cost per successful answer and the fastest-five human-time baseline is often a reasonable empirical summary, but its exponent points the opposite way from the search-efficiency story: most fitted exponents are below $1$, including those of the strongest high-coverage base models, and the bootstrap $90\%$ CIs lie entirely below $1$ for three of the four strongest base models (Claude Opus 4.6, GPT-5.4 (xhigh), and Gemini 3.1 Pro Preview), with the fourth (GPT-5.2 (high)) just barely crossing $1$ at the upper end. The finding is robust across our sensitivity checks: it survives changes to the success-rate filter (\Cref{rem:threshold_sensitivity}), to the choice of $k$ in the human-time baseline (\Cref{tab:top_k_sensitivity}), and to bootstrap resampling of the included problems. Within the proof-search framing of \Cref{sec:complexitymodel}, this means the AI systems we tested scale \emph{better} than the strong human Project Euler solvers as problems get harder.

Hypothesis~P fares better. For most of the well-covered models the binned success-vs-difficulty curve is close to a straight line in semi-log coordinates with the one-parameter ansatz $\log p_{\text{success}} = c \cdot t_{\text{human}}$, giving median bin-level $R^2 = 0.92$ across 22 models, with 13 of 22 configurations exceeding $R^2 = 0.90$. As an additional sanity check (\Cref{rem:hypp_two_param}), the unconstrained two-parameter fit on the same bins lands near $y(0\,\text{h}) = 100\%$ for most configurations even though the data does not include a bin at $t = 0$ to anchor it, indicating that the strict one-parameter form is not too far off. We read this together as mild empirical support for the underlying constant-deviation picture, with the caveats from \Cref{sec:interpreting_hyp_p}: the fitted half-life moves with the choice of bin count, and Project Euler problems --- which often reward a single mathematical reduction --- are not the cleanest test bed for a model of long reasoning chains.

The METR-style analysis gives the most directly readable summary. Under the same fastest-five human baseline, the best current $50\%$ Project Euler horizons in this dataset lie between roughly $2.5$ and $4.3$ hours, and agentic wrappers are associated with materially larger horizons than their underlying base models, though the comparison is observational and confounded by differing attempt counts and scaffolding choices. The current absolute horizons are short relative to the full range of human Project Euler solve times, so what visibly limits these systems on this benchmark is reliability rather than search efficiency: the models hit $50\%$ success at human work-times of only a few hours, well short of the time strong human solvers can sustain on a single problem.

Read together, the three analyses give a coherent picture. Hypothesis~T says that token cost per successful answer grows \emph{sublinearly} in $t_{\text{human}}$, contradicting the search-efficiency prediction of $b > 1$. Hypothesis~P says that success probability decays exponentially in $t_{\text{human}}$, so what limits these systems on harder problems is reliability rather than effort per success. The METR analysis converts that decay into a model-by-model human-time at which the success curve crosses $50\%$ (and $80\%$), giving the order-of-magnitude horizons reported above.

\section*{AI Methods Statement}

AI systems were used in several parts of this project. GPT-5 was used in earlier stages of the data-collection code. Claude Opus 4.1 was used to draft early prose and summarize existing material. Claude Code (Sonnet 4.5) was used to help write some analysis and plotting scripts. For the present revision, OpenAI Codex (GPT-5-based) was used to review the analysis code, generate additional analysis scripts and paper figures, and draft proposed manuscript revisions; Claude Code (Opus 4.7) was used to review the manuscript and the analysis code, to produce supplementary validation analyses together with the top-$k$ sensitivity and bootstrap-CI outputs, and to draft the corresponding prose additions. All mathematical claims, dataset descriptions, fitted values, and final wording were checked and curated by the human authors.

\section*{Acknowledgments}

We thank Timothy Gowers for proposing the original search-complexity model and the participants of the Lorentz Center workshop for valuable discussions. We thank Michiel van der Meer for bringing the literature on inverse test-time scaling to our attention.  We are particularly grateful to Jasper Dekoninck for sharing the Project Euler evaluation data and for detailed feedback and methodological input on this paper, including the overthinking hypothesis discussed in \Cref{sec:speculation}. Johannes Schmitt was supported by the Swiss National Science Foundation grant 10009122, ``Beyond Benchmark Scores: Analyzing AI Reasoning on Research-Level Mathematics''. 

\appendix

\section{Hypothesis~T: Full Per-Model Results}
\label{app:hypothesis_t_plots}

\Cref{tab:hypothesis_t_summary} reports the fitted parameters $a$, $b$, $R^2$, the bootstrap $90\%$ CI on $b$, the included-problem count, and the overall success rate for all 25 model configurations. The per-model log-log plots that follow show generated-token cost per successful answer against the fastest-five human-time baseline, restricted to model-problem pairs with empirical success rate at least $50\%$.

{
\footnotesize\setlength{\tabcolsep}{4pt}
\begin{longtable}{lrrrrrr}
\caption{Power-law fits for Hypothesis T. Here $y = a x^b$, where $y$ is the generated-token cost per successful answer, defined as total generated tokens across all attempts on a model-problem pair divided by the number of successful attempts, and $x$ is the geometric mean of the five fastest human solve times for the corresponding problem, in seconds. Only model-problem pairs with at least 50\% success enter the fit; the \textbf{Included} column reports the number of such pairs out of the problems covered by the model, and \textbf{Success rate} is the overall attempt-level success rate across all covered problems, not just the included ones. The \textbf{b 90\% CI} column is a percentile bootstrap interval from $10^3$ resamples of the included model-problem cells. The coefficient $a$ is unit-dependent (it absorbs the choice of seconds for $x$), so we interpret only the exponent $b$ and the fit quality $R^2$.}\label{tab:hypothesis_t_summary}\\
\toprule
\textbf{Model} & \textbf{$a$} & \textbf{$b$} & \textbf{$b$ 90\% CI} & \textbf{$R^2$} & \textbf{Included} & \textbf{Success rate} \\
\midrule
\endfirsthead
\toprule
\textbf{Model} & \textbf{$a$} & \textbf{$b$} & \textbf{$b$ 90\% CI} & \textbf{$R^2$} & \textbf{Included} & \textbf{Success rate} \\
\midrule
\endhead
\midrule
\multicolumn{7}{r}{Continued on next page} \\
\midrule
\endfoot
\bottomrule
\endlastfoot
Claude Opus 4.6 & 2.4e+02 & 0.722 & [0.59, 0.88] & 0.564 & 41/47 & 89.3\% \\
Grok 4 Fast (Euler Agent 1) & 9.8e+02 & 0.727 & [0.51, 0.97] & 0.604 & 17/19 & 89.0\% \\
GPT-5.4 (xhigh) & 3.0e+02 & 0.644 & [0.49, 0.80] & 0.541 & 47/50 & 88.8\% \\
Gemini 3.1 Pro Preview & 1.4e+03 & 0.467 & [0.36, 0.60] & 0.468 & 46/50 & 88.3\% \\
Grok 4 Fast (older Euler agent) & 1.3e+01 & 1.258 & [1.07, 1.58] & 0.599 & 17/21 & 83.1\% \\
GPT-5.2 (high) & 8.0e+01 & 0.813 & [0.61, 1.03] & 0.528 & 38/44 & 80.7\% \\
GPT-5 (Euler Agent) & 2.1e+02 & 0.932 & [0.51, 1.22] & 0.376 & 19/24 & 79.2\% \\
Grok 4 Fast (Euler Agent 2) & 2.6e+01 & 1.176 & [0.76, 1.45] & 0.609 & 17/24 & 68.8\% \\
Grok 4 Fast (Euler Agent 3) & 1.1e+01 & 1.327 & [0.96, 1.76] & 0.589 & 16/24 & 66.7\% \\
GLM 5.1 & 2.9e+02 & 0.775 & [0.49, 0.99] & 0.443 & 31/49 & 66.3\% \\
GPT-5.1 (high) & 3.1e+03 & 0.360 & [0.16, 0.60] & 0.398 & 22/33 & 65.2\% \\
Gemini 3 Pro (preview) & 1.2e+03 & 0.480 & [0.25, 0.64] & 0.326 & 29/42 & 59.5\% \\
Gemini 3 Flash & 3.0e+03 & 0.351 & [0.11, 0.52] & 0.238 & 28/42 & 58.9\% \\
Kimi K2.5 (Think) & 2.8e+03 & 0.435 & [0.26, 0.64] & 0.356 & 28/47 & 58.2\% \\
GPT-5 (high) & 2.4e+03 & 0.400 & [0.15, 0.62] & 0.383 & 19/28 & 58.0\% \\
Grok 4 (Euler Agent) & 3.8e+00 & 1.526 & [0.99, 2.14] & 0.489 & 17/24 & 54.8\% \\
Grok 4 Fast (August solver scaffold) & 3.4e+02 & 0.689 & [0.40, 1.05] & 0.508 & 12/24 & 54.4\% \\
Kimi K2 Thinking & 1.1e+04 & 0.224 & [-0.03, 0.48] & 0.086 & 21/40 & 50.0\% \\
DeepSeek-v3.2 (Think) & 4.7e+03 & 0.325 & [0.06, 0.52] & 0.144 & 23/43 & 48.3\% \\
o4-mini (high) & 3.9e+03 & 0.325 & [0.17, 0.66] & 0.158 & 14/28 & 48.2\% \\
Grok 4 & 2.2e+03 & 0.443 & [0.02, 0.84] & 0.220 & 16/31 & 45.5\% \\
Grok 4 Fast & 1.1e+03 & 0.505 & [0.10, 1.10] & 0.230 & 13/28 & 44.6\% \\
GPT-5 (Euler Agent 2) & 1.1e+00 & 1.601 & [0.36, 7.68] & 0.560 & 4/9 & 44.4\% \\
Grok 4.1 Fast & 4.2e+02 & 0.671 & [0.47, 0.94] & 0.391 & 21/44 & 42.0\% \\
Gemini 2.5 Pro & 2.1e+03 & 0.531 & [-0.83, 1.44] & 0.235 & 6/28 & 15.2\% \\
\end{longtable}
}

\includepdf[pages=-,pagecommand={\thispagestyle{plain}}]{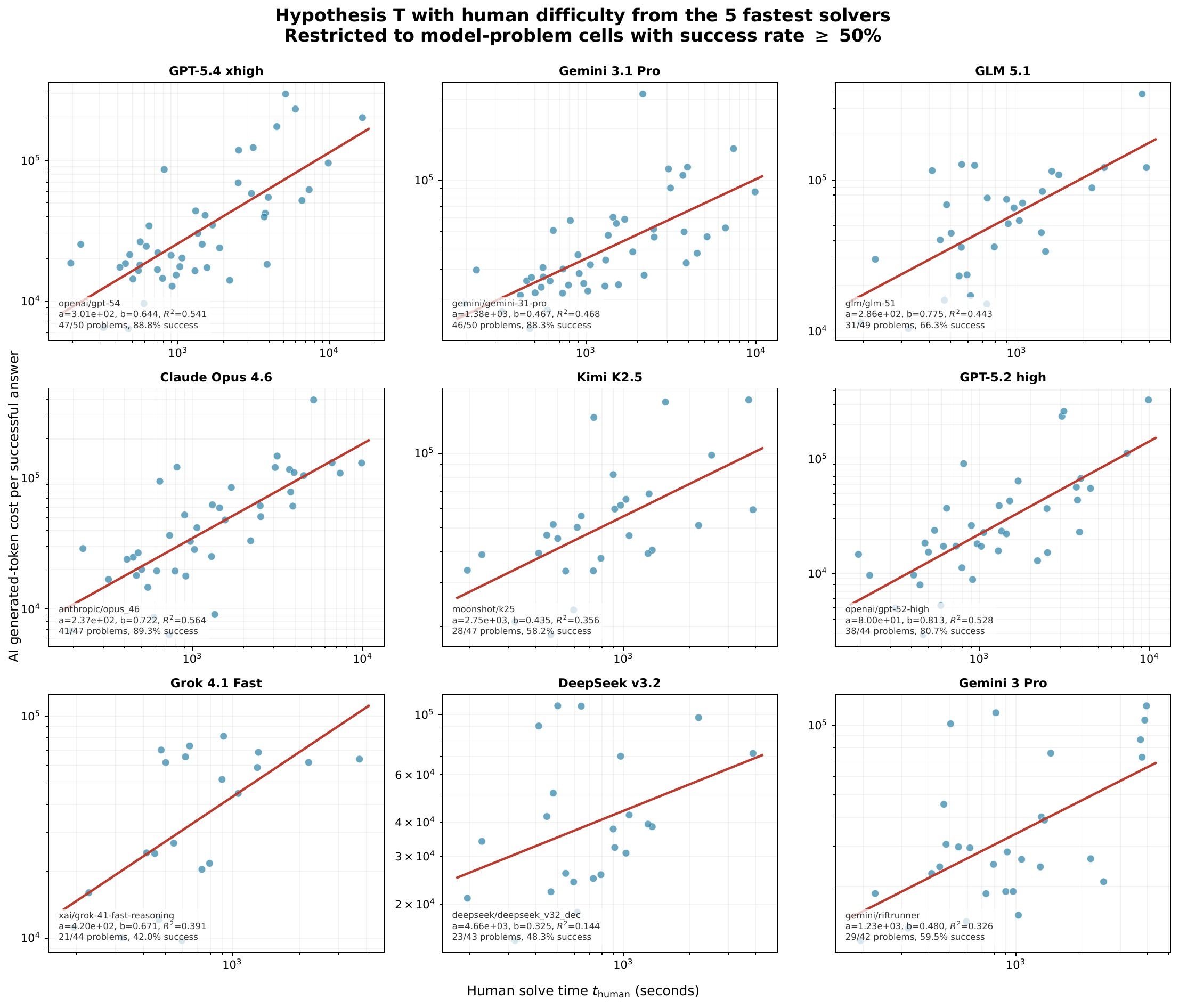}

\section{Hypothesis~P: Full Per-Model Results and Bin Sensitivity}
\label{app:hypothesis_p}

\begin{figure}[!htbp]
\centering
\includegraphics[width=\textwidth]{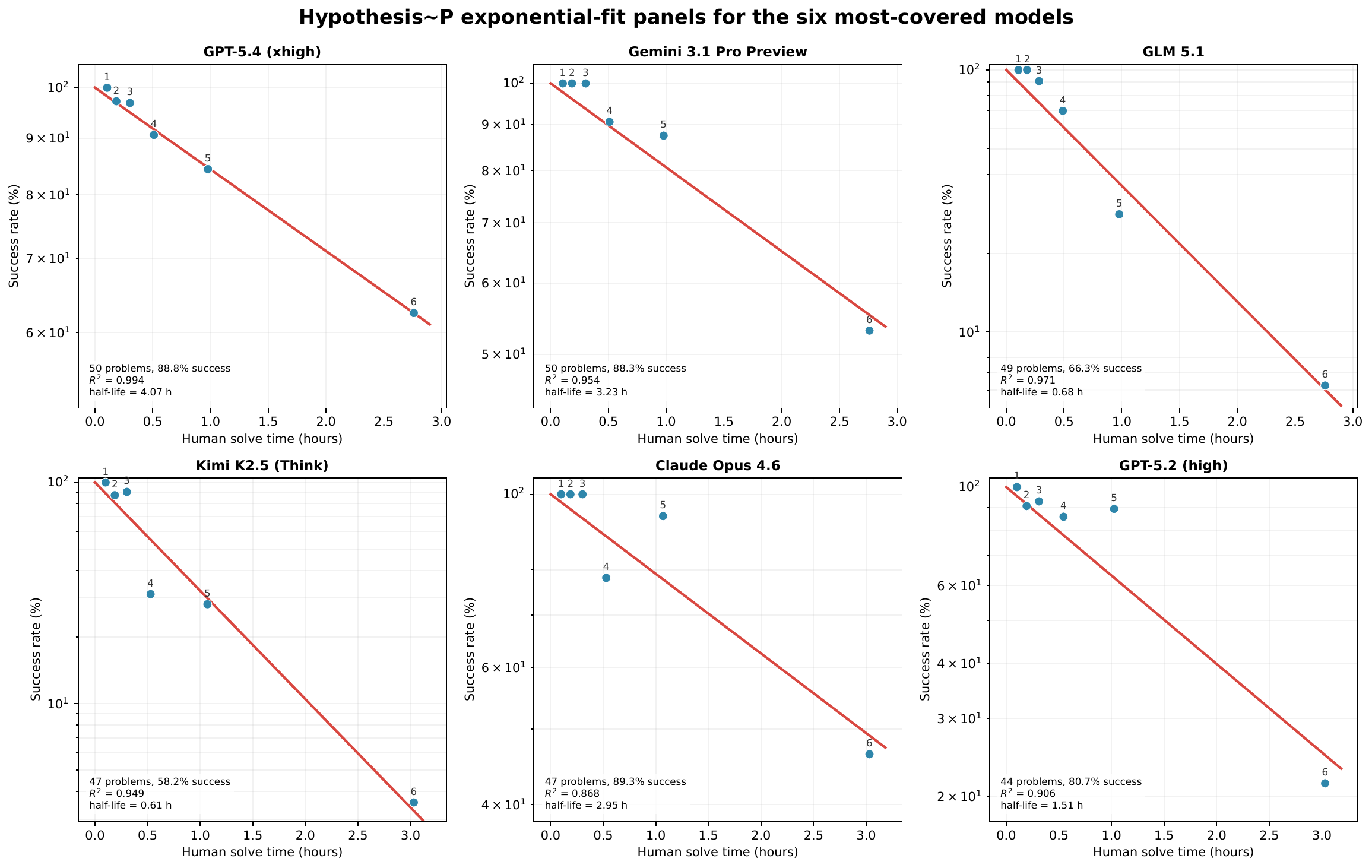}
\caption{Hypothesis~P semi-log exponential-fit panels for the six models with the largest number of analyzed problems. Bin averages are computed after sorting problems by the fastest-five human baseline; bins with $0\%$ success are omitted from the fit, and the fit is constrained to pass through $(t_{\text{human}}, p_{\text{success}}) = (0, 1)$.}
\label{fig:hypothesis_p_top6}
\end{figure}

\Cref{fig:hypothesis_p_top6} shows the Hypothesis~P fit panels for the six most-covered model configurations, and \Cref{tab:hypothesis_p_summary} extends the summary to all 22 model configurations with at least 24 covered problems.

{
\tiny\setlength{\tabcolsep}{3pt}
\begin{longtable}{lrrrrr}
\caption{Six-bin Hypothesis~P fit summary for models with at least 24 analyzed problems. Problems are sorted by the fastest-five human baseline and averaged within bins; the fit is the one-parameter ansatz $\log p_{\text{success}} = c \cdot t_{\text{human}}$ (equivalently $p_{\text{success}} = e^{c\, t_{\text{human}}}$), with $0\%$ bins dropped from the regression. The half-life is $-\log 2 / c$ in hours. The last column reports the METR-style $50\%$ horizon on the same human baseline for comparison.}\label{tab:hypothesis_p_summary}\\
\toprule
\textbf{Model} & \textbf{Problems} & \textbf{Success rate} & \textbf{$R^2$} & \textbf{Half-life (h)} & \textbf{METR $h_{50}$ (h)} \\
\midrule
\endfirsthead
\toprule
\textbf{Model} & \textbf{Problems} & \textbf{Success rate} & \textbf{$R^2$} & \textbf{Half-life (h)} & \textbf{METR $h_{50}$ (h)} \\
\midrule
\endhead
\midrule
\multicolumn{6}{r}{Continued on next page} \\
\midrule
\endfoot
\bottomrule
\endlastfoot
GPT-5.4 (xhigh) & 50 & 88.8\% & 0.994 & 4.07 & 4.28 \\
Gemini 3.1 Pro Preview & 50 & 88.3\% & 0.954 & 3.23 & 2.57 \\
GLM 5.1 & 49 & 66.3\% & 0.971 & 0.68 & 0.72 \\
Claude Opus 4.6 & 47 & 89.3\% & 0.868 & 2.95 & 3.38 \\
Kimi K2.5 (Think) & 47 & 58.2\% & 0.949 & 0.61 & 0.56 \\
GPT-5.2 (high) & 44 & 80.7\% & 0.906 & 1.51 & 1.82 \\
Grok 4.1 Fast & 44 & 42.0\% & 0.866 & 0.28 & 0.30 \\
DeepSeek-v3.2 (Think) & 43 & 48.3\% & 0.843 & 0.34 & 0.37 \\
Gemini 3 Pro (preview) & 42 & 59.5\% & 0.781 & 0.73 & 0.60 \\
Gemini 3 Flash & 42 & 58.9\% & 0.930 & 0.63 & 0.58 \\
Kimi K2 Thinking & 40 & 50.0\% & 0.905 & 0.24 & 0.37 \\
GPT-5.1 (high) & 33 & 65.2\% & 0.472 & 0.87 & 0.68 \\
Grok 4 & 31 & 45.5\% & 0.703 & 0.38 & 0.33 \\
GPT-5 (high) & 28 & 58.0\% & 0.650 & 0.81 & 0.56 \\
o4-mini (high) & 28 & 48.2\% & 0.967 & 0.31 & 0.37 \\
Grok 4 Fast & 28 & 44.6\% & 0.958 & 0.29 & 0.32 \\
Gemini 2.5 Pro & 28 & 15.2\% & 0.974 & 0.12 & 0.09 \\
GPT-5 (Euler Agent) & 24 & 79.2\% & 0.965 & 1.69 & 1.44 \\
Grok 4 Fast (Euler Agent 2) & 24 & 68.8\% & 0.981 & 1.07 & 0.92 \\
Grok 4 Fast (Euler Agent 3) & 24 & 66.7\% & 0.994 & 0.82 & 0.81 \\
Grok 4 (Euler Agent) & 24 & 54.8\% & 0.887 & 1.16 & 0.70 \\
Grok 4 Fast (August solver scaffold) & 24 & 54.4\% & 0.873 & 0.49 & 0.37 \\
\end{longtable}
}

\subsection{Sensitivity to bin count}

The Hypothesis~P summaries in \Cref{sec:hypothesis_p_results} use six equal-size quantile bins of the covered problems sorted by $t_{\text{human}}$. Two implementation choices interact with that bin count. First, bins with $0\%$ success are dropped from the semi-log fit, since $\log 0 = -\infty$, rather than replaced by an arbitrary floor. Second, increasing the bin count improves resolution along $t_{\text{human}}$ but reduces per-bin sample size and so the stability of the bin success rate.

\begin{figure}[!htbp]
\centering
\includegraphics[width=\textwidth]{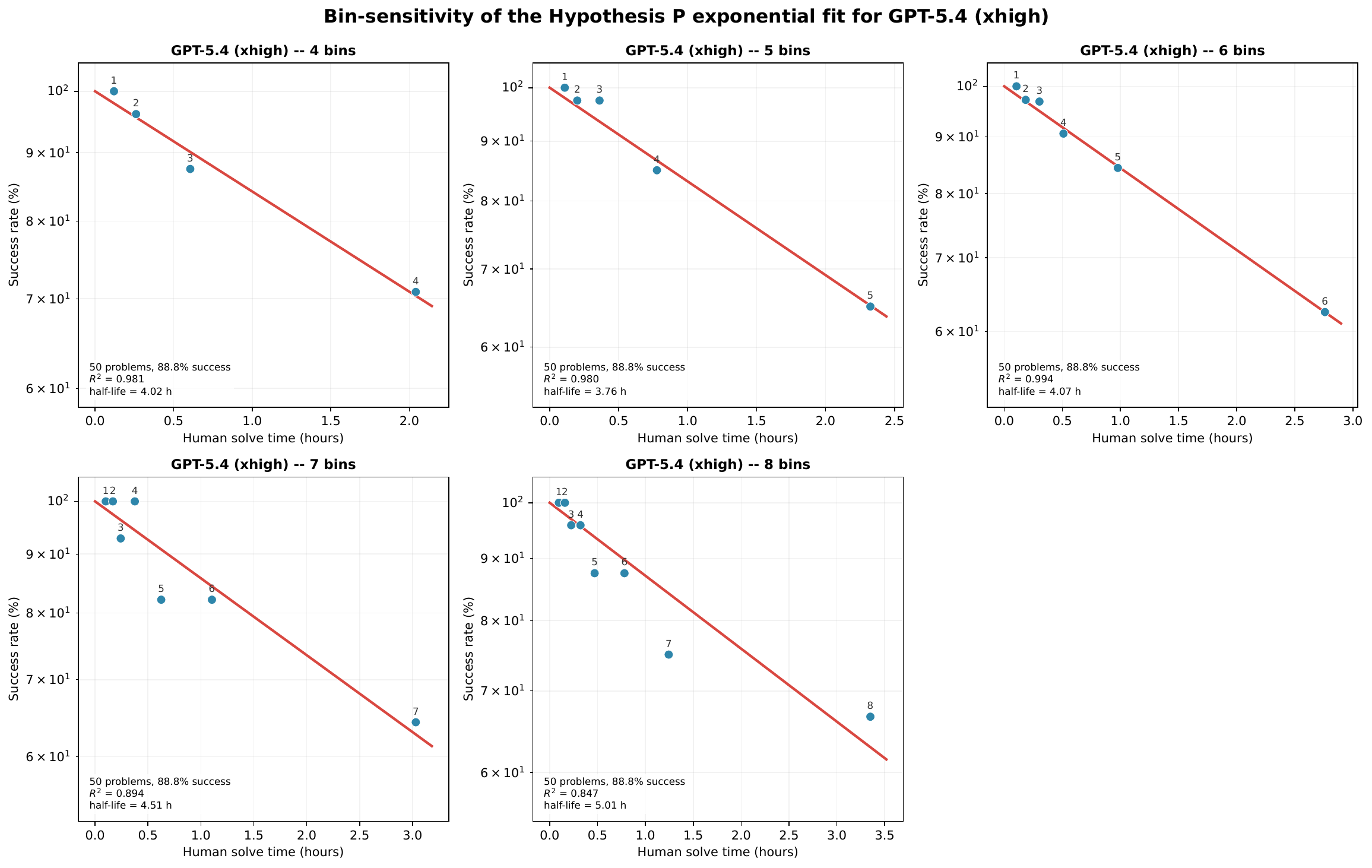}
\caption{Sensitivity of the Hypothesis~P semi-log exponential fit to the choice of bin count, for GPT-5.4 (xhigh). Each panel uses the same 50 covered problems, partitioned into 4--8 equal-size quantile bins along $t_{\text{human}}$, fit under the constrained one-parameter ansatz \eqref{eq:hyp_p_linear}. The fitted half-life moves modestly with bin count, while bin-level $R^2$ degrades visibly past 6 bins as per-bin sample size shrinks.}
\label{fig:hypothesis_p_gpt54_bin_sens}
\end{figure}

\Cref{fig:hypothesis_p_gpt54_bin_sens} shows GPT-5.4 (xhigh), the best-covered configuration in our dataset, fit at bin counts $\{4, 5, 6, 7, 8\}$ under the one-parameter ansatz \eqref{eq:hyp_p_linear}. The fitted half-life moves over a range of about $1.3$~hours, from $3.76$~hours at 5 bins to $5.01$~hours at 8 bins. Bin-level $R^2$ exceeds $0.98$ for $\{4, 5, 6\}$ bins; with $\{7, 8\}$ bins the per-bin sample sizes shrink to at most $7$ problems each (out of 50), and the fit quality drops to $R^2 \in \{0.85, 0.89\}$. We choose six bins in the main text as a compromise between resolution along $t_{\text{human}}$ and per-bin sample size; no choice in this range materially changes the qualitative picture.

\section{METR-Style Full Per-Model Results}
\label{app:metr_plots}

\Cref{tab:metr_summary} reports the fitted $h_{50}$ and $h_{80}$ horizons together with their bootstrap $90\%$ CIs for the non-agent base models with adequate coverage and the two base/agent comparison pairs (GPT-5 (high) vs.\ GPT-5 (Euler Agent), and Grok 4 Fast vs.\ Grok 4 Fast (Euler Agent 1)); the remaining agentic scaffolds, which share underlying release dates with the listed base models, are omitted from the table. The per-model logistic panels that follow visualize the underlying fits.

{
\footnotesize\setlength{\tabcolsep}{4pt}
\begin{longtable}{lrrrrrr}
\caption{METR-style Project Euler time horizons under the fastest-five human baseline. All non-agent base models with adequate coverage are shown, together with two base/agent comparison pairs: GPT-5 (high) vs GPT-5 (Euler Agent), and Grok 4 Fast vs Grok 4 Fast (Euler Agent 1). \textbf{Success rate} is the overall attempt-level success rate across the problems covered by the model. \textbf{90\% CI} columns are percentile bootstrap intervals from $10^3$ resamples of the covered problems; entries shown as $[\ell, {>}100]$ indicate a bootstrap upper bound above $100$\,h, where the logistic fit on small problem counts produces near-flat curves for some resamples.}\label{tab:metr_summary}\\
\toprule
\textbf{Model} & \textbf{Problems} & \textbf{Success rate} & \textbf{$h_{50}$ (h)} & \textbf{$h_{50}$ 90\% CI} & \textbf{$h_{80}$ (h)} & \textbf{$h_{80}$ 90\% CI} \\
\midrule
\endfirsthead
\toprule
\textbf{Model} & \textbf{Problems} & \textbf{Success rate} & \textbf{$h_{50}$ (h)} & \textbf{$h_{50}$ 90\% CI} & \textbf{$h_{80}$ (h)} & \textbf{$h_{80}$ 90\% CI} \\
\midrule
\endhead
\midrule
\multicolumn{7}{r}{Continued on next page} \\
\midrule
\endfoot
\bottomrule
\endlastfoot
GPT-5.4 (xhigh) & 50 & 88.8\% & 4.28 & [2.74, 10.14] & 1.35 & [0.89, 2.27] \\
Claude Opus 4.6 & 47 & 89.3\% & 3.38 & [1.77, 9.25] & 1.27 & [0.72, 2.55] \\
Gemini 3.1 Pro Preview & 50 & 88.3\% & 2.57 & [1.75, 5.03] & 1.30 & [0.97, 2.17] \\
GPT-5.2 (high) & 44 & 80.7\% & 1.82 & [1.21, 3.05] & 0.74 & [0.49, 1.15] \\
Grok 4 Fast (Euler Agent 1) & 19 & 89.0\% & 1.73 & [0.82, ${>}100$] & 0.68 & [0.36, ${>}100$] \\
GPT-5 (Euler Agent) & 24 & 79.2\% & 1.44 & [0.78, 2.50] & 0.82 & [0.48, 1.93] \\
GLM 5.1 & 49 & 66.3\% & 0.72 & [0.56, 0.95] & 0.41 & [0.31, 0.56] \\
GPT-5.1 (high) & 33 & 65.2\% & 0.68 & [0.46, 1.01] & 0.35 & [0.22, 0.58] \\
Gemini 3 Pro (preview) & 42 & 59.5\% & 0.60 & [0.48, 0.77] & 0.27 & [0.20, 0.37] \\
Gemini 3 Flash & 42 & 58.9\% & 0.58 & [0.44, 0.80] & 0.28 & [0.21, 0.40] \\
GPT-5 (high) & 28 & 58.0\% & 0.56 & [0.38, 0.84] & 0.23 & [0.13, 0.38] \\
Kimi K2.5 (Think) & 47 & 58.2\% & 0.56 & [0.42, 0.74] & 0.28 & [0.22, 0.38] \\
o4-mini (high) & 28 & 48.2\% & 0.37 & [0.26, 0.54] & 0.18 & [0.13, 0.28] \\
DeepSeek-v3.2 (Think) & 43 & 48.3\% & 0.37 & [0.28, 0.48] & 0.20 & [0.14, 0.27] \\
Kimi K2 Thinking & 40 & 50.0\% & 0.37 & [0.30, 0.44] & 0.23 & [0.17, 0.30] \\
Grok 4 & 31 & 45.5\% & 0.33 & [0.24, 0.44] & 0.16 & [0.10, 0.24] \\
Grok 4 Fast & 28 & 44.6\% & 0.32 & [0.24, 0.46] & 0.19 & [0.14, 0.27] \\
Grok 4.1 Fast & 44 & 42.0\% & 0.30 & [0.24, 0.38] & 0.16 & [0.13, 0.21] \\
Gemini 2.5 Pro & 28 & 15.2\% & 0.09 & [0.03, 0.15] & 0.04 & [0.01, 0.10] \\
\end{longtable}
}

\includepdf[pages=-,pagecommand={\thispagestyle{plain}}]{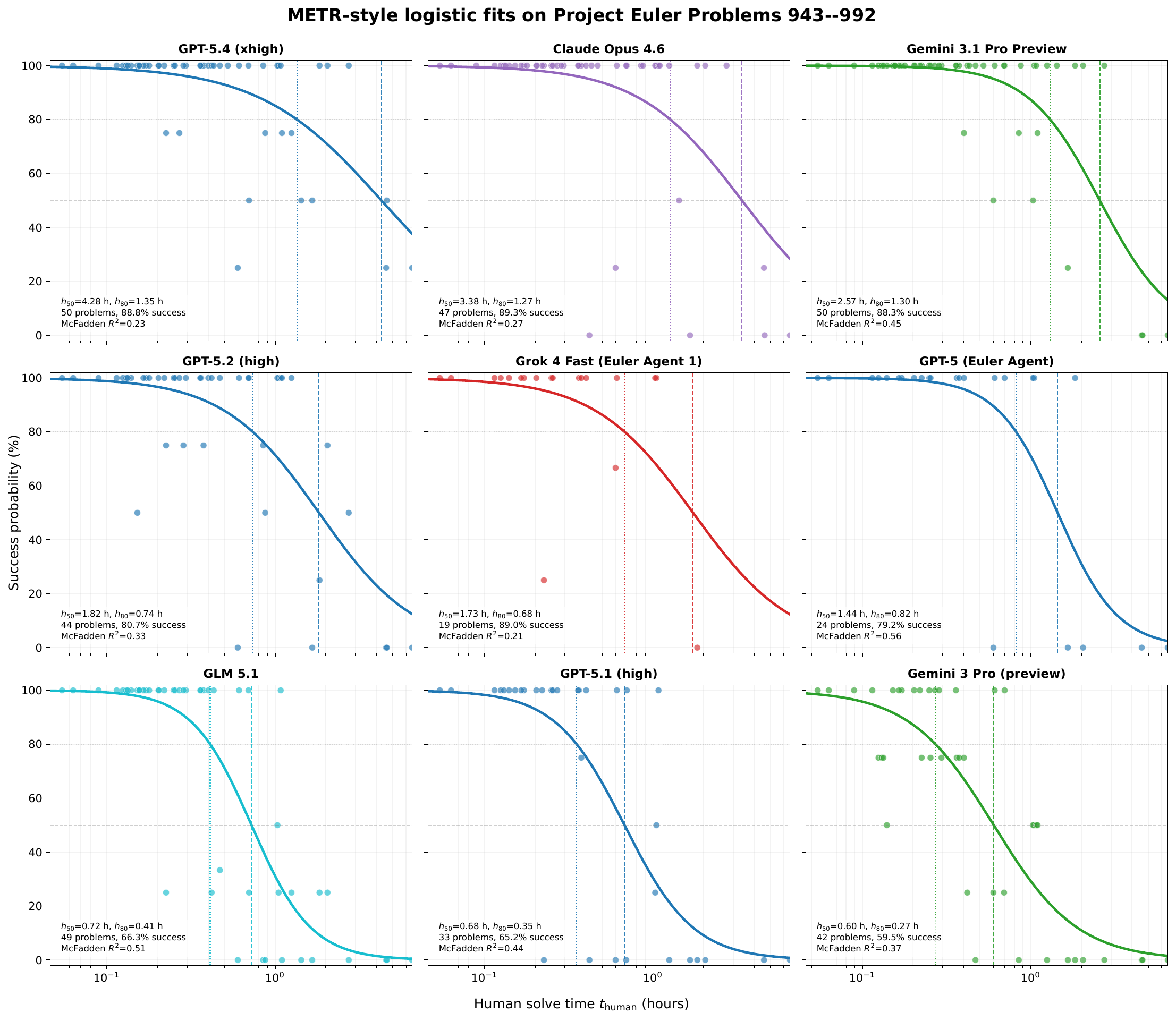}

\bibliographystyle{alpha}
\bibliography{main}

\end{document}